\documentclass[journal]{IEEEtran}

\usepackage{xcolor}
\usepackage{xspace}
\usepackage{MnSymbol}
\usepackage{makecell}
\usepackage{multirow}

\ifCLASSINFOpdf
\usepackage[pdftex]{graphicx}
\graphicspath{ {Figures/Anomalies/} }
\DeclareGraphicsExtensions{.pdf,.jpeg,.png}
\else
\fi
\usepackage{amsmath}
\usepackage[linesnumbered]{algorithm2e}

\ifCLASSOPTIONcompsoc
\usepackage[caption=false,font=normalsize,labelfont=sf,textfont=sf]{subfig}
\else
  \usepackage{floatrow}
  \usepackage[caption=false,font=footnotesize]{subfig}
  \usepackage{caption}
\fi

    \usepackage{hyperref}
 
    \hypersetup{
        colorlinks = true,
        linkcolor=red,
        citecolor=red,
        urlcolor=blue,
        linktocpage, 
        breaklinks=false
    }

\begin{document}
	
	    \bstctlcite{bstctl:etal, bstctl:nodash, bstctl:simpurl}
		%
		\title{Classification of Anomalies in Telecommunication Network KPI Time Series}
		%
		%
		%
		
		\author{\IEEEauthorblockN{Korantin~Bordeau--Aubert \IEEEauthorrefmark{1}, 
                                      Justin Whatley \IEEEauthorrefmark{2}, 
                                      Sylvain Nadeau \IEEEauthorrefmark{2}, 
                                      Tristan~Glatard \IEEEauthorrefmark{1}, 
                                      Brigitte Jaumard \IEEEauthorrefmark{1}
                                      }
  
  \IEEEauthorblockA{\IEEEauthorrefmark{1}Department of Computer Science and Software Engineering, Concordia University, Montreal, Quebec, Canada.}

\IEEEauthorblockA{\IEEEauthorrefmark{2} EXFO Inc., Montreal, QC, Canada.}

  }

	\maketitle

	
	\begin{abstract}
		The increasing complexity and scale of telecommunication networks have led to a growing interest in automated anomaly detection systems. However, the classification of anomalies detected on network Key Performance Indicators (KPI) has received less attention, resulting in a lack of information about anomaly characteristics and classification processes. To address this gap, this paper proposes a modular anomaly classification framework. The framework assumes separate entities for the anomaly classifier and the detector, allowing for a distinct treatment of anomaly detection and classification tasks on time series. The objectives of this study are (1) to develop a time series simulator that generates synthetic time series resembling real-world network KPI behavior, (2) to build a detection model to identify anomalies in the time series, (3) to build classification models that accurately categorize detected anomalies into predefined classes (4) to evaluate the classification framework performance on simulated and real-world network KPI time series. This study has demonstrated the good performance of the anomaly classification models trained on simulated anomalies when applied to real-world network time series data.
	\end{abstract}
	

	%
	\IEEEpeerreviewmaketitle

\section{Introduction}
\label{sec:introduction}
\IEEEPARstart{A}{nomaly} detection in network KPI time series has gained significant attention in recent years due to the increasing complexity and scale of the systems. Communication networks generate massive amounts of time series data capturing various activities and behaviors. The detection of anomalies in time series data is crucial in traffic analysis and performance monitoring or Key Performance Indicators (KPI). Compared to anomaly detection, classification of anomalies detected on network KPI has received relatively less attention in research, giving less information on the anomalies characteristics and classification processes. Exploiting classification techniques to identify and categorize anomalies within network KPI time series data remains an open area for exploration. Further research is needed to find new approaches and methodologies to accurately detect and classify anomalies, enhancing network KPI monitoring capabilities.
	
This article addresses the problem of effectively simulating anomalies on KPI and assessing anomaly classification models on both simulated and real-world time series datasets. In this paper, we separate the detector and classifiers as two parts within the anomaly classification framework. The detector identifies the anomalies in the time series while the classifiers categorize the time series into their corresponding classes or types. This separation allows for a modular approach between the detection and the classification.
	
This paper introduces an anomaly classification framework. This framework includes a network KPI time series simulator, a detection model, classification models. The detection and classification models are evaluated on both simulated and real-world time series datasets. The simulator generates simulated time series with anomalies to mimic real-world network KPI systems. This simulator provides time series to train and test the detection and classification models. The detection model is built to identify the anomalies within the simulated and real-world time series. Additionally, we aim to build classification models to accurately classify the anomalies in their respective classes. Finally, we evaluate the performance and generalization capabilities of these classification models by testing them on both simulated datasets, generated by our time series simulator, and real datasets, obtained from network KPI.

The paper is organized as follows. Section~\ref{sec:literature} provides a comprehensive review of existing literature on network KPI anomaly definition, anomaly classification models, and evaluation methodologies. Section~\ref{sec:method} presents our method for the anomaly classification framework, detailing the simulation, the detection and classification models. Finally, Section~\ref{sec:experiments} presents the experimental setup, including the datasets used and the performance of the models on both simulated and real-world time series datasets, followed by concluding remarks in Section~\ref{sec:conclusion}.


\section{Background}
\label{sec:literature}
This background section aims to provide an overview of the key concepts, methodologies, and advancements in the field of network anomaly definition, simulation, detection, and classification.

Defining anomalies is the first step in the creation of a simulator and classification model. Network KPI anomalies encompasses a wide range of definitions and interpretations, reflecting the diverse nature of anomalies that can occur in network systems. Choi \textit{et al.} \cite{choi2021deep} identified 3 major anomaly types in time series: point, contextual, and collective anomalies. Point anomalies refer to a data point or sequence that exhibits a sudden and significant deviation from the normal range. These anomalies are typically caused by sensor errors or abnormal system operations and are detected by comparing values against predefined upper and lower control limits. Contextual anomalies present a challenge for detection as they do not deviate from the normal range based on predefined limits. This type of anomaly is characterized by a group of points that do not contains extreme values and modify the shape of the signal. Collective anomalies represent a set of data points that gradually display a different pattern from normal data over time. While individual values within this type may not appear anomalous, their collective behavior deviates from the expected baseline. Detecting collective anomalies requires examining long-term contexts to identify deviations from the expected pattern. Foorthuis \cite{foo2018}, \cite{foorthuis2021nature} proposed the following five fundamental data-oriented dimensions for describing types and subtypes of anomalies: data type, cardinality of relationship, anomaly level, data structure, and data distribution. These dimensions contribute to the classification and characterization of anomalies in network data. The data type dimension differentiates between quantitative, qualitative, and mixed attributes. The cardinality of relationship dimension distinguishes between univariate and multivariate relationships among attributes. The anomaly level dimension classifies anomalies as either atomic (individual low-level cases) or aggregate (groups or collective structures). The data structure dimension considers different structural formats, such as graphs and time series, which host specific anomaly subtypes. Lastly, the data distribution dimension focuses on the collection and pattern of attribute values in the data space, providing additional descriptive and delineating capabilities for anomaly classification. We focused our research on quantitative and univariate time series with atomic and aggregate anomalies. The anomaly types used in this paper are the Single point, Temporary change, Level shift and variation change anomalies.

Network anomaly detection techniques aim to identify abnormal patterns or events within network data. Traditional approaches include statistical methods, rule-based systems, and expert systems, which rely on predefined thresholds or rules to flag deviations. Forecasting is one of the most used techniques for anomaly detection in time series, as it enables real-time analysis. A popular forecasting approach is based on the ARIMA \cite{box1970distribution} (Autoregressive Integrated Moving Average) model, which combines autoregressive (AR), integrative (I), and moving average (MA) components to capture the underlying patterns and characteristics of time series. When evaluating and forecasting stationary time series data, ARIMA models are useful because they produce predictions for the future by considering both historical data and prediction errors. Deep learning techniques for time series forecasting, including Recurrent Neural Networks (RNN) \cite{elman1990finding, madan2018predicting} and Convolutional Neural Networks (CNN) \cite{lecun1998gradient, borovykh2017conditional}, have also shown promise in capturing temporal dependencies and spatial patterns for improved anomaly detection, as explained by Choi \textit{et al.} \cite{choi2021deep}. Furthermore, Malhotra \textit{et al.} \cite{malhotra2015long}  demonstrated that stacked LSTM networks can effectively learn temporal patterns and detect anomalies without prior knowledge of pattern duration. The LSTM-AD approach demonstrates promising results on real-world datasets, outperforming or matching RNN-AD and indicating the robustness of LSTM-based models in capturing both short-term and long-term dependencies in normal time series behavior. More recently, the need to combine CNN and RNN approaches in time series detection has been seen as essential \cite{xue2019evolving}, as it allows for the simultaneous extraction of spatial features and temporal dependencies. This combination enables more comprehensive and accurate analysis of time-varying patterns in the data. Lea \textit{et al.} \cite{lea2016temporal} proposed the Temporal Convolutional Network (TCN) as a unified approach between RNN and CNN architectures. The TCN demonstrates comparable or better performance than other models on various datasets, and it offers the advantage of faster training. Moreover Bai \textit{et al.} \cite{bai2018empirical} highlights the superiority of TCNs over generic recurrent architectures, such as LSTM and Gated Recurrent Units (GRU) \cite{cho2014learning}, in various sequence modeling tasks. The TCN model, incorporating dilations, residual connections, and causal convolutions, consistently outperforms recurrent architectures. The study also challenges the notion that recurrent networks have an inherent advantage in preserving long-range dependencies, as TCN demonstrate comparable or even longer memory capabilities. Overall, the results suggest that convolutional networks, with their simplicity and clarity, should be considered a strong foundation and a powerful toolkit for sequence modeling. We used the TCN model for anomaly detection and classification purpose in this paper.

Network anomaly classification focuses on categorizing detected anomalies into meaningful classes. Classification models utilize features extracted from network data to differentiate between different types of anomalies, such as intrusion attempts, Denial of Service (DoS) attacks, or network performance issues. Faouzi \cite{faouzi2022time} reviewed the algorithm and implementation of Time series classification (TSC). The study described various approaches such as the Nearest Neighbors (NN) \cite{bagnall2014experimental} classifier combined with the Dynamic Time Warping (DTW) distance. 
Extensive comparisons conducted by Bagnall \textit{et al.} \cite{bagnall2017great} confirmed that DTW outperforms other distance measures. Another approach was with tree-based algorithm, such as the  Time series Forest (TSF) \cite{deng2013time}. Cabello \textit{et al.} \cite{random_forest} introduced the Supervised Time series Forest (STSF) as an evolution of the TSF \cite{deng2013time}, an efficient algorithm for interval-based time series classification on high-dimensional datasets. STSF leverages multiple time series representations, a supervised search strategy, and a feature ranking metric to reduce computational overhead while identifying highly discriminatory interval features for interpretable classification outcomes. The paper stated STSF achieved comparable accuracy to state-of-the-art methods in time series classification but with significantly faster processing times, enabling classification of large datasets with long series. Anomaly classification models also witnessed a transition from traditional rule-based and statistical approaches to more advanced techniques such as machine learning, deep learning, and ensemble methods, enabling improved accuracy and generalization capabilities in detecting and categorizing anomalies. Zhao \textit{et al.} \cite{zhao2017convolutional} suggested that a CNN-based method for time series classification outperforms competing baseline methods in terms of classification accuracy and noise tolerance. By automatically discovering and extracting the internal structure of input time series, the CNN generates deep features that improve classification performance. The study also discusses the significance of three CNN parameters: convolutional filter size, pooling method, and the number of convolution filters. However, limitations are acknowledged, including the time-consuming nature of CNN training due to parameter experimentation and the fixed length requirement for time series during training and testing. Ongoing research aims to address these limitations by designing optimal parameters and exploring a new network architecture that combines CNN with RNN. Ismail F. \textit{et al.} \cite{ismail2020inceptiontime} introduced InceptionTime, a CNN-based classifier achieving state-of-the-art performance UCR \cite{UCRArchive} archive datasets. This study also discusses the importance of the parameters such as the depth and the number of filters. Even though numerous studies on time series classification have been carried out, more research and analysis are still needed in the specific area of network anomaly classification. Studies concentrating solely on network anomaly classification in time series data are noticeably scarce, with little study devoted to this particular field. Although network anomaly detection is extremely important, research into and development of classification models specifically suited for network KPI time series data is still in its early stages.

    \section{Anomaly Simulation and Classification Framework}
    \label{sec:method}
\begin{figure}[ht!]
	\centering
	\includegraphics[width=0.95\linewidth]{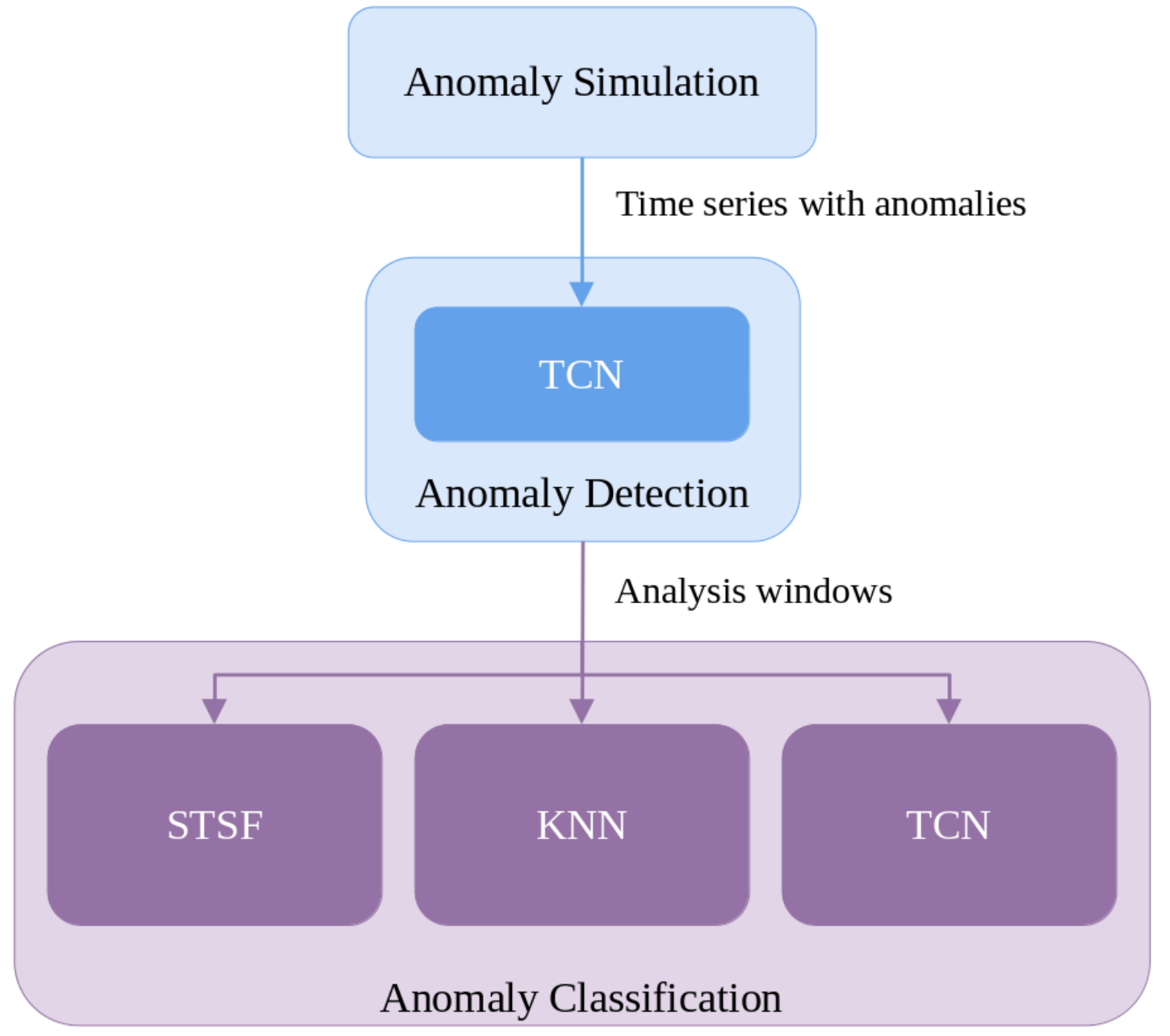}
	\caption{Anomaly classification framework}
	\label{fig:framework}
\end{figure}

	Our framework separates anomaly detection from anomaly classification (Fig. \ref{fig:framework}). This section describes the simulation method to generate anomalies, as well as the detection and classification models used in the framework.

\subsection{Anomaly Simulation}
\label{subsec:simulation}

    \begin{table}[ht!]
        \scriptsize
        \centering
        \begin{tabular}{l|l}
            n & number of anomalies to inject\\
            X & initial time series\\
            $\tilde{X}$ & time series with anomalies\\
            $\bar{X}$ & simulated noise time series\\
            $\hat{X}$ & final time series with anomalies and noise\\
            $\check{X}$ & predicted time series\\
            A & amplitude of the sine signals\\
            T & seasonality period of the sine signals in minutes\\
            $t_s$ & sampling period\\
            $\mu$ & mean of the sine signals\\
            $\bar\mu$ & mean of the normal distribution for the noise generation\\
            $\bar\sigma$ & standard deviation of the normal distribution for the noise generation\\
            $\sigma$ & noise level of the time series\\
            $A_d$ & daily amplitude of the time series\\
            $\lambda$ & length of the anomalies\\
            $\alpha$ & strength of the anomaly\\
            $i_w$ & first index of an anomaly window\\
            $\delta$ & confidence interval to determine a point as anomalous\\
            $m$ & margin period of the analysis windows\\
            $T_i$, $U_i$, $R_i$ & time series features (seasonal, trend and residual)\\

        \end{tabular}
    
        \caption{Notations}
        \label{table:notations}
        
    \end{table}

\begin{figure}[H]
		\centering
		\includegraphics[width=\linewidth]{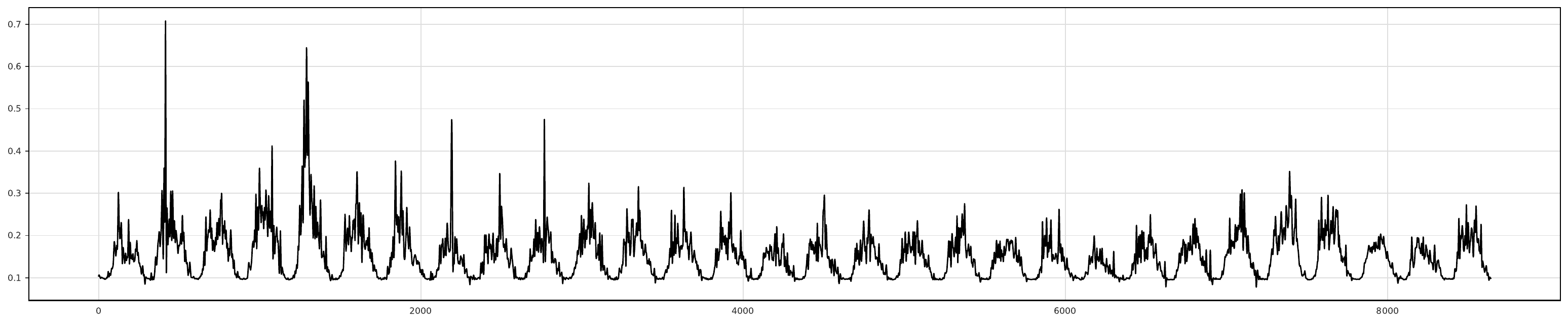}
		\caption{Example of REAL time series after processing}
		\label{fig:real_ts}
	\end{figure}

    Anomaly simulation aims to generate network Key Performance Indicators (KPI) time series. Network KPIs are used by companies to track network performance from metrics such as packet loss or latency. These indicators gives an overview of the network usage and health. In this paper, we focus on latency, which is typically modeled with 3 seasonality components (daily, weekly, and monthly seasonality), a trend, and a noise component (Fig.~\ref{fig:real_ts}). The latency is commonly measured as follows: (1) a packet is sent from point A to B in the network, (2) the packet is sent back from B to A, (3) steps (1) and (2) are repeated for a given number of packets, (4) the latency is obtained by averaging the packet round-trip time between A and B. We denote the data sets respectively as SIM for the time series resulting from anomaly simulation and REAL for the real time series data sets.
    
    \subsubsection{Signal Generation}
    \label{subsubsec:signal-generation}
    
\begin{figure}[H]
	        \centering
	        \includegraphics[width=\linewidth]{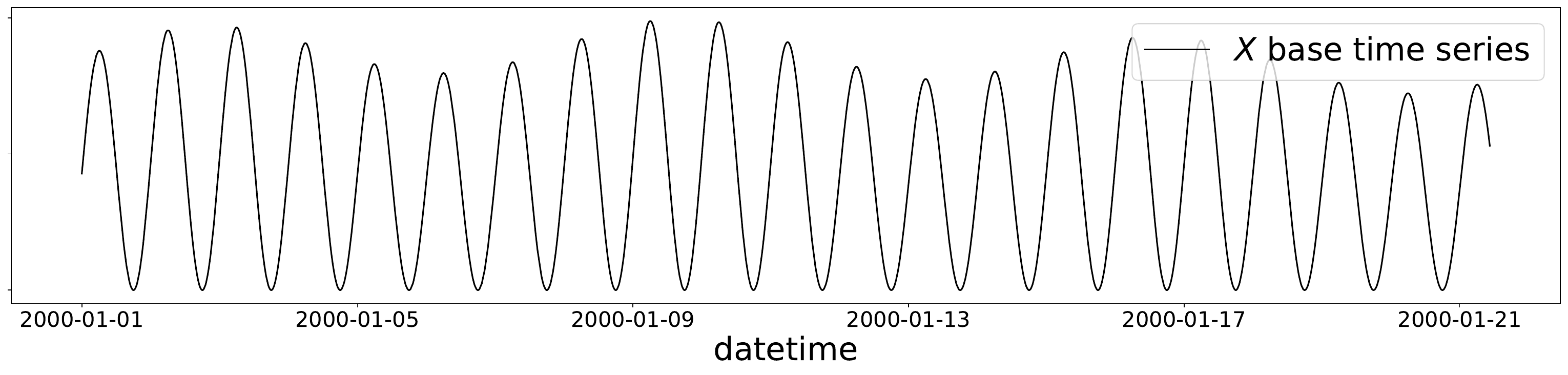}
	        \caption{$X$ base time series.}
	        \label{fig:sim_base_ts}
        \end{figure}

        We define an \emph{anomaly window} as a sub-sequence of the original time series containing exactly one anomaly. We also define $n$ the number of anomalies to inject in the SIM time series. We generate $n$ anomaly windows, as follows: (1) we randomly select the anomaly class to generate based on anomaly proportions set as a parameters; (2) we set the anomaly window size to $2Y$ points where $Y$ is a random variable following the uniform distribution $\mathcal{U}(l_{min}, l_{max})$. We empirically set the limits $l_{min}$, $l_{max}$ depending on the anomaly type. The size of the SIM time series corresponds to the sum of each anomaly window size. We generate the SIM base time series using 3 seasonality components modeled with 3 sine signals with daily, weekly and monthly periodicity:
        \[
            X(t) = \prod\limits_{s=0}^{2} \left(A_s \times \sin\left( \dfrac{2\pi}{T_s}t\right) + \mu_s \right),
        \]
        where $A_s$ is the amplitude: $A = (A_0, A_1, A_2) = (0.5, 0.1, 0.05)$, $A$ represent in reality the average delay in $\mu s$; $T_s$ is the seasonality period expressed in minutes: $T = (T_0, T_1, T_2) = (1440, 10080, 40320)$; $\mu_s$ is the mean of the sine signal: $\mu = (\mu_0, \mu_1, \mu_2) = (0.5, 0.9, 0.95)$; and $t$ is expressed in minutes. The amplitudes and means add up to $1$, to keep the base time series between $0$ and $1 \mu s$ which corresponds to typical latency values. The different amplitudes are set according to observations on the REAL data sets. The shape of this base time series is consistent with common observations of network traffic that usually include these pseudo-periods without noise. The time series $X$ is then uniformly sampled with the period $t_s$ (Fig. \ref{fig:sim_base_ts}).
        
    \subsubsection{Anomaly Injection}
    \label{subsubsec:anomaly-injection}

        We define each anomaly type with 2 parameters:
        \begin{enumerate}
            \item The anomaly length $\lambda$, defined as the number of points modified during injection in the time series. To select the length of each anomaly, we randomly select a starting index $i_a$ in $[i_{w}, i_{w} + 2Y - e]$ where $i_w$ is the index of the first point in the anomaly window and $e$ corresponds to the minimum margin $e = 5$ that ensures a minimal size for each class of anomalies. $\lambda$ is then randomly selected in $[i_{a}, i_{w} + 2Y - e]$. 
            \item The strength $\alpha$, which depends on the daily amplitude $A_d$: \[A_d = \max_{i \in d} X(i) - \min_{i \in d} X(i),\] 
        where $d$ is the daily range such that $i_a \in d$. $d$ corresponds to a daily period of length $\frac{1440}{t_s}$ minutes. $\alpha$ is then defined as:
        \[\alpha = A_d \times Z,\]
        where $Z$ is a random variable following the uniform distribution $\mathcal{U}(0.5, 0.7)$. 
        \end{enumerate}
        
\begin{figure}[H]
            \centering
            \includegraphics[width=\linewidth]{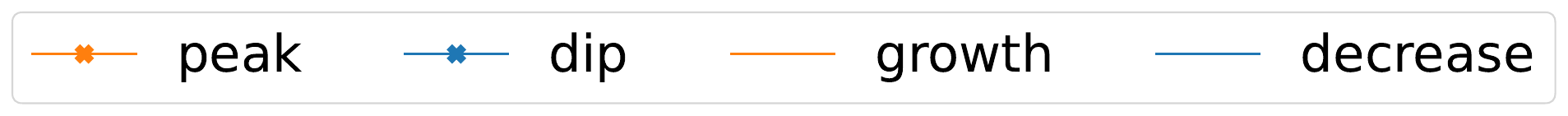}
            \subfloat[Single point]{
                \includegraphics[width=\linewidth]{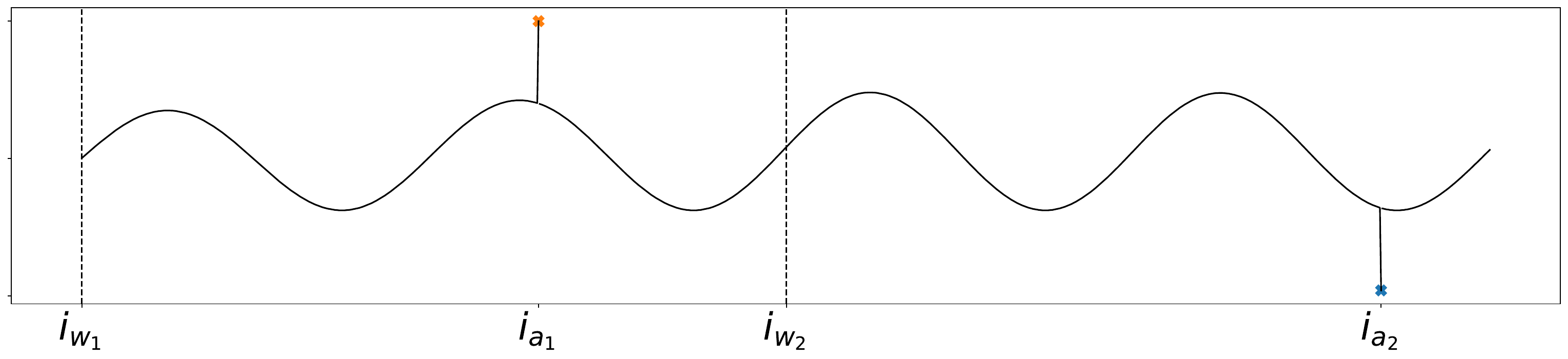}
                \label{fig:single_point}
            }
            \\
            \subfloat[Temporary change]{
                \includegraphics[width=\linewidth]{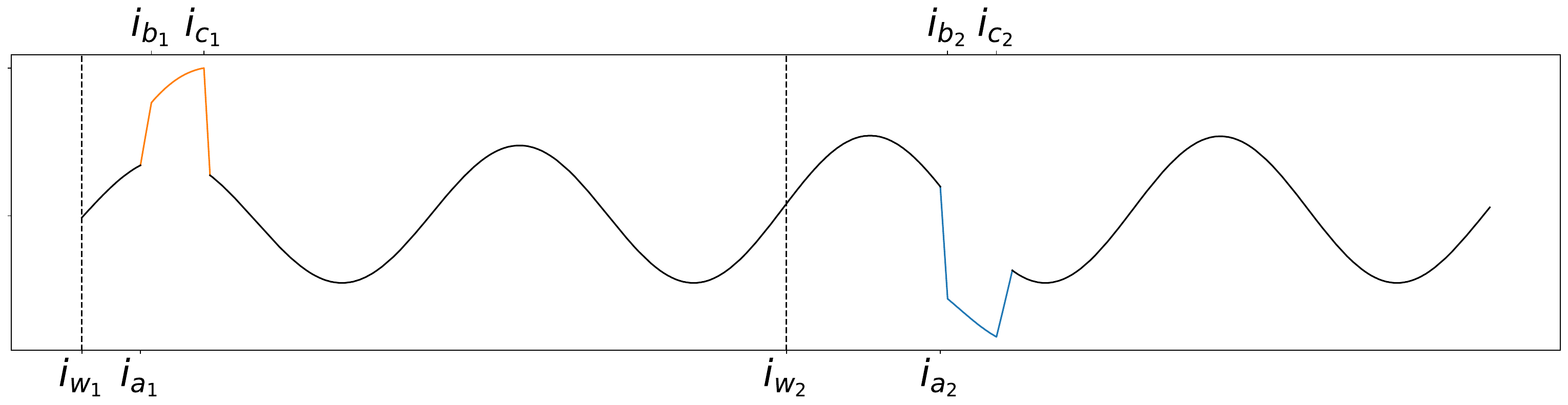}
                \label{fig:tmp_change}
            }
            \\
            \subfloat[Level shift]{
                \includegraphics[width=\linewidth]{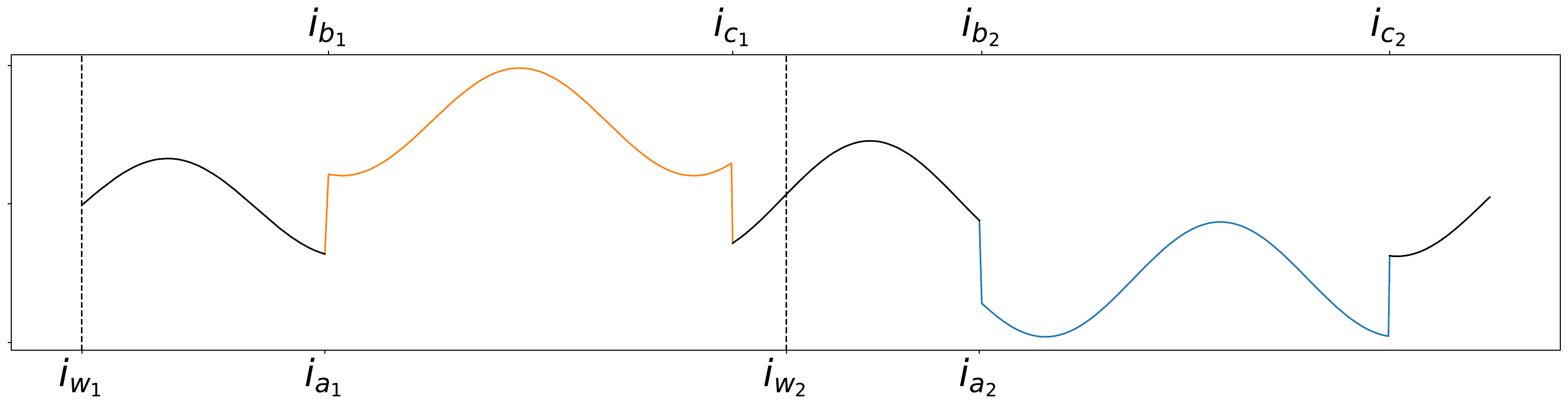}
                \label{fig:level_shift}
            }
            \\
            \subfloat[Variation change]{
                \includegraphics[width=\linewidth]{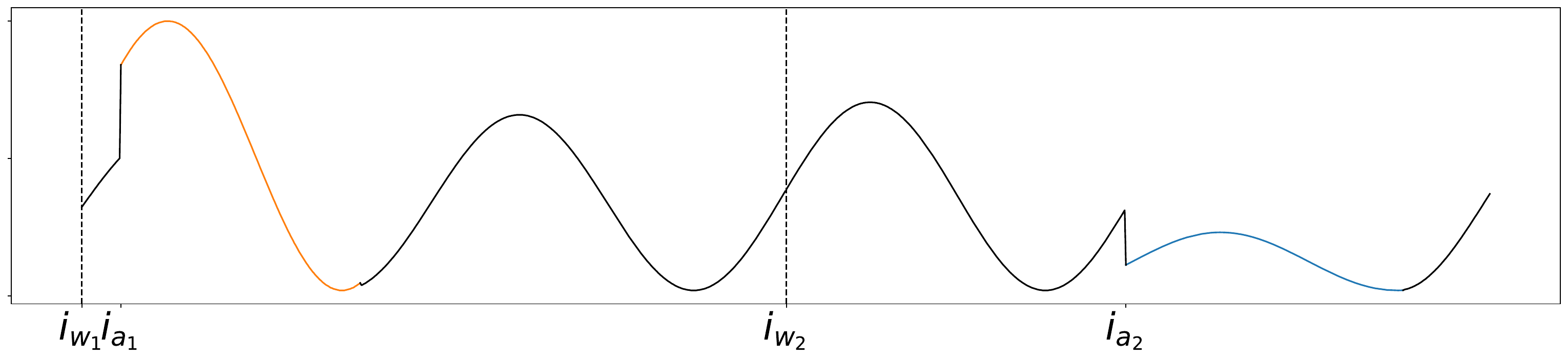}
                \label{fig:variation_change}
            }
            \caption{Types of anomalies.}
            \label{fig:anomalies_type}
        \end{figure}

        We denote by $\tilde{X}$ the time series containing a simulated anomaly. The 4 simulated types of anomalies are the following:
        
        \paragraph{Single point (peak or dip)} This anomaly type has a length $\lambda$ equal to $1$ point and $Y$ follows the uniform distribution $\mathcal{U}(120, 480)$. The anomaly window is defined for $i$ in $[i_w, i_w+2Y]$ as:
        
        \[
            \tilde X(i) = X(i) \pm
            \begin{cases}
                \alpha & \text{if } i=i_a\\
                0 & \text{otherwise},
            \end{cases}
        \]
        In this formula, peaks are generated by the addition and dips by the subtraction. See illustration in Figure \ref{fig:single_point}.

        \paragraph{Temporary change (growth or decrease)} The length of this anomaly type varies in $[3, i_w + 2Y]$ and $Y$ follows the uniform distribution $\mathcal{U}(240, 960)$.  The anomaly is described as a growth or decrease that progressively reverts to the original signal:
        \[
            \tilde X(i) = X(i) \pm
            \begin{cases}
                (i-i_a)\dfrac{\alpha_1}{i_b-i_a} 
                & \text{if } i \in [i_a, i_b] \\
                (i-i_b)\dfrac{\alpha_2-\alpha_1}{i_c-i_b} + \alpha_1
                & \text{if } i \in [i_b, i_c] \\
                (i-i_a-\lambda)\dfrac{\alpha_2}{i_c-i_a-\lambda} 
                & \text{if } i \in [i_c, i_a+\lambda]\\
                0 & \mathrm{otherwise,}
            \end{cases}
        \]
        where $i_b$ is randomly selected in [$i_a$, $i_a+\lambda/2$], $i_c$ is randomly selected in [$i_b$, $i_a+\lambda$], and $\alpha_1, \alpha_2$ are either set to $\alpha$ and between $[0.4, \alpha]$ or vice versa. See illustration in Figure \ref{fig:tmp_change}. 
        	
        \paragraph{Level shift (growth or decrease)} The length of this anomaly type varies in $[3, i_w + 2Y]$ and $Y$ follows the uniform distribution $\mathcal{U}(1440, 2160)$. This type of anomaly is a variant of the previous one where $\alpha_1=\alpha_2$. It is defined by a change of the time series trend:
        \[
            \tilde X(i) = X(i) \pm
            \begin{cases}
                (i-i_a)\dfrac{\alpha}{i_b-i_a} 
                & \text{if } i \in [i_a, i_b] \\
                \alpha
                & \text{if } i \in [i_b, i_c] \\
                (i-i_a-\lambda)\dfrac{\alpha}{i_c-i_a-\lambda} 
                & \text{if } i \in [i_c, i_a+\lambda] \\
                0 & \text{otherwise,}
            \end{cases}
        \]
        where $i_b$ is randomly selected in [$i_a$, $i_a+\lambda/2$], and $i_c$ is randomly selected in [$i_b$, $i_a+\lambda$].
        See illustration in Figure \ref{fig:level_shift}.

        \paragraph{Variation change (growth or decrease)} The length of this anomaly type varies in $[3, i_w + 2Y]$ and $Y$ follows the uniform distribution $\mathcal{U}(1440, 2160)$. This type of anomaly amplifies or reduces the amplitude of a time series while preserving its period:
        \[
            \tilde X(i) = X(i) \pm
            \begin{cases}
                X(i) \alpha
                & \text{if } i \in[i_a,i_a+\lambda]\\
                0 & \text{otherwise},
            \end{cases}
        \]
        See illustration in Figure \ref{fig:variation_change}, where the signals only include daily seasonality (weekly and monthly seasonalities are omitted) to simplify presentation. The simulator supports 4 anomaly types where each type can be separated into 2 sub-classes (growth or decrease).
        
    \subsubsection{Data Preparation and Noise Injection}
    \label{subsubsec:data_preparation_noise_injection}
        
        \paragraph{Scaling} We apply the MinMaxScaler\footnote{\url{https://scikit-learn.org/stable/modules/generated/sklearn.preprocessing.MinMaxScaler.html}} function from the scikit-learn library to scale the time series to $[0.02, 1]$. The minimum is fixed to $0.02$, which is the range observed in the real dataset. Indeed, real latency values can never reach exactly 0. This scaler assure the time series are between the same range for both the detection and classification for the SIM and REAL data sets.        
        
\begin{figure}[H]
	        \centering
	        \includegraphics[width=\linewidth]{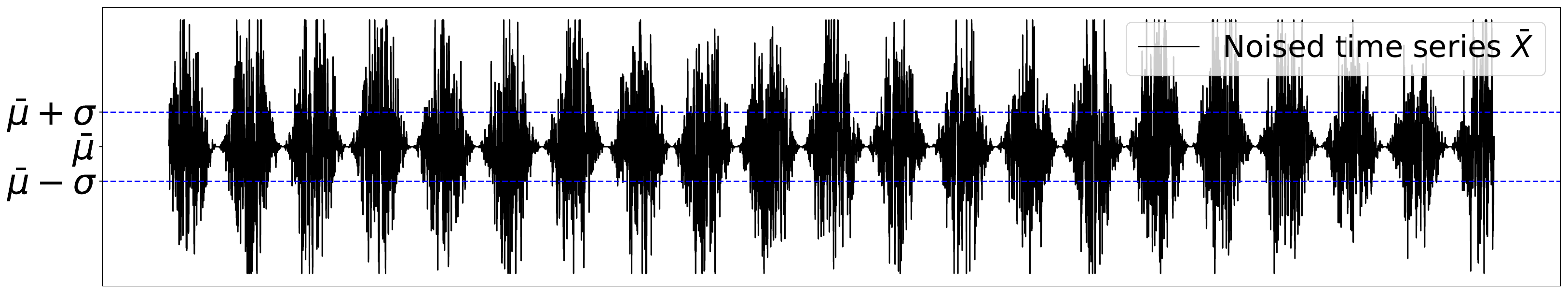}
	        \caption{Noised time series $\bar X$}
	        \label{fig:sim_ts_n}
        \end{figure}

        \paragraph{Noise Injection} We create a noise time series $\bar X$ with equal length to $\tilde{X}$. This noise is defined by a Gaussian multiplicative white noise:
        \[
            \bar X (i) = X(i) \times
            \begin{cases}
                \min (W, 4\bar\sigma)  & \text{if } W \geq 0 \\
                \max (W, -4\bar\sigma) & \textbf{}{otherwise},
            \end{cases}
        \]
        where $W$ is a random variable following the normal distribution $\mathcal{N}(\bar\mu, \bar\sigma)$, and $\bar\sigma$ is the standard deviation of the simulated Gaussian white noise time series, see illustration in Figure~\ref{fig:sim_ts_n}. We remove the extreme values to evaluate the anomalies injected by the simulator and not those resulting from the noise. We separate the noise level $\sigma$ of the time series from the standard deviation $\bar\sigma$ of the noised time series. Indeed, as we multiplied the Gaussian white noise time series by $X(i)$, the noise level $\sigma$ extracted from the final time series is not equal to the noise level $\bar\sigma$ we used during the noise generation. Indeed, due to the seasonal shape of the initial time series, we obtained $\bar\sigma = c\times\sigma$ with the constant $c=2.31$. 
        
        The Single point and Temporary change anomalies aren't affected by the noise as they are part of the same frequencies. Indeed, the noise is already contained in these types of anomalies. 
        
        We denote $\hat X$ the time series resulting from the scaling and noise injection:
        \[
            \hat X (i) = \tilde{X}(i) +
            \begin{cases}
                0 & \mathrm{if}\ \tilde{X}(i)\in \text{Single point}\\
                0 & \mathrm{if}\ \tilde{X}(i)\in \text{Temporary change}\\
                \bar X (i) & \mathrm{otherwise},
            \end{cases}
        \]
        
\begin{figure}[H]
        \centering
        \includegraphics[width=\linewidth]{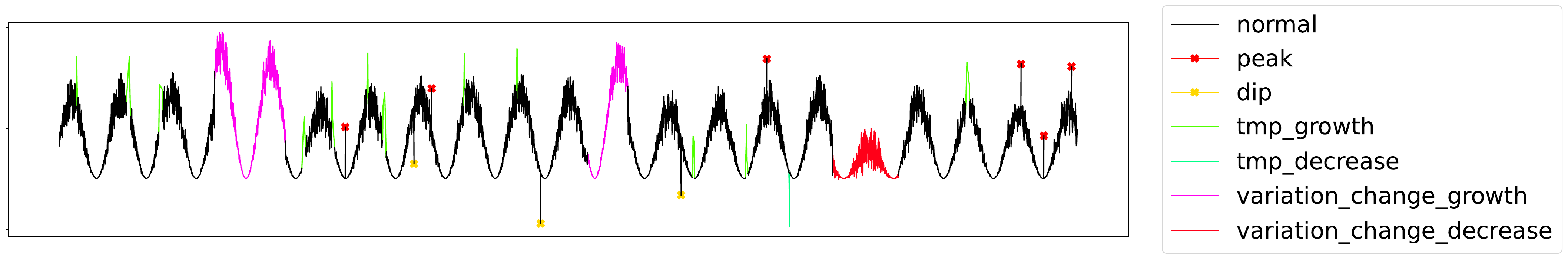}
        \caption{Final simulated time series (SIM $\hat{X}$)}
        \label{fig:sim_ts}
    \end{figure}

    The resulting time series $\hat{X}$ is used in the SIM data sets, as illustrated in Figure \ref{fig:sim_ts}. $\hat{X}$ also corresponds to the processed REAL data sets. The parameters used to generate the anomalies are the noise level, the proportion of each type of anomalies, and the sampling period $t_s$. The sampling period is an important parameter of the simulation. For instance, a Single point anomaly simulated for a given sampling period may be interpreted as a Temporary change for another sampling period.
    
\subsection{Anomaly Detection}
\label{subsec:anomaly_detection}

    To evaluate the anomaly \emph{classification} model, we build an anomaly \emph{detection} model using a Temporal Convolutional Network (TCN~\cite{bai2018empirical}), a common approach to time series prediction~\cite{torres_deep_2021}. We used the TCN implementation of the darts library\footnote{\url{https://unit8co.github.io/darts/generated\_api/darts.models.forecasting.tcn\_model.html}}, varying the parameters according to Table~\ref{table:tcn_detection_param}.

    \begin{table}[ht!]
        \centering
        \begin{tabular}{|l|l|}
            \hline
            Parameter & Values tested\\
            \hline
            Kernel size & 2, 3, 4, 5, 6, 7\\\hline
            Number of filters & 4, 5, 6, 7, 8\\\hline
            Dilation base & 3, 4, 5\\\hline
            Input chunk length & $(1440 / t_s)\times 8$\\\hline
            Output chunk length & $(1440 / t_s)\times 7$\\
            \hline
        \end{tabular}
        \caption{Parameter combinations tested for the TCN time series parameter.}
        \label{table:tcn_detection_param}    
    \end{table}

    \paragraph{Detection model} The TCN detection model is trained for 10 epochs with the Adam optimizer. We limit the number of residual block to $2$ to reduce the training time. This model is composed of ReLU activation functions with a $\beta$-Negative log likelihood loss \cite{seitzer2022pitfalls} from the Gaussian likelihood function\footnote{\url{https://unit8co.github.io/darts/generated\_api/darts.utils.likelihood\_models.html\#darts.utils.likelihood\_models.GaussianLikelihood}}. The Gaussian likelihood function provides boundaries for the predicted time series with a $95\%$ confidence interval $\delta$. The detection model gives a weekly prediction from the previous 8 days.
    
\begin{figure}[H]
        \centering
        \includegraphics[width=\linewidth]{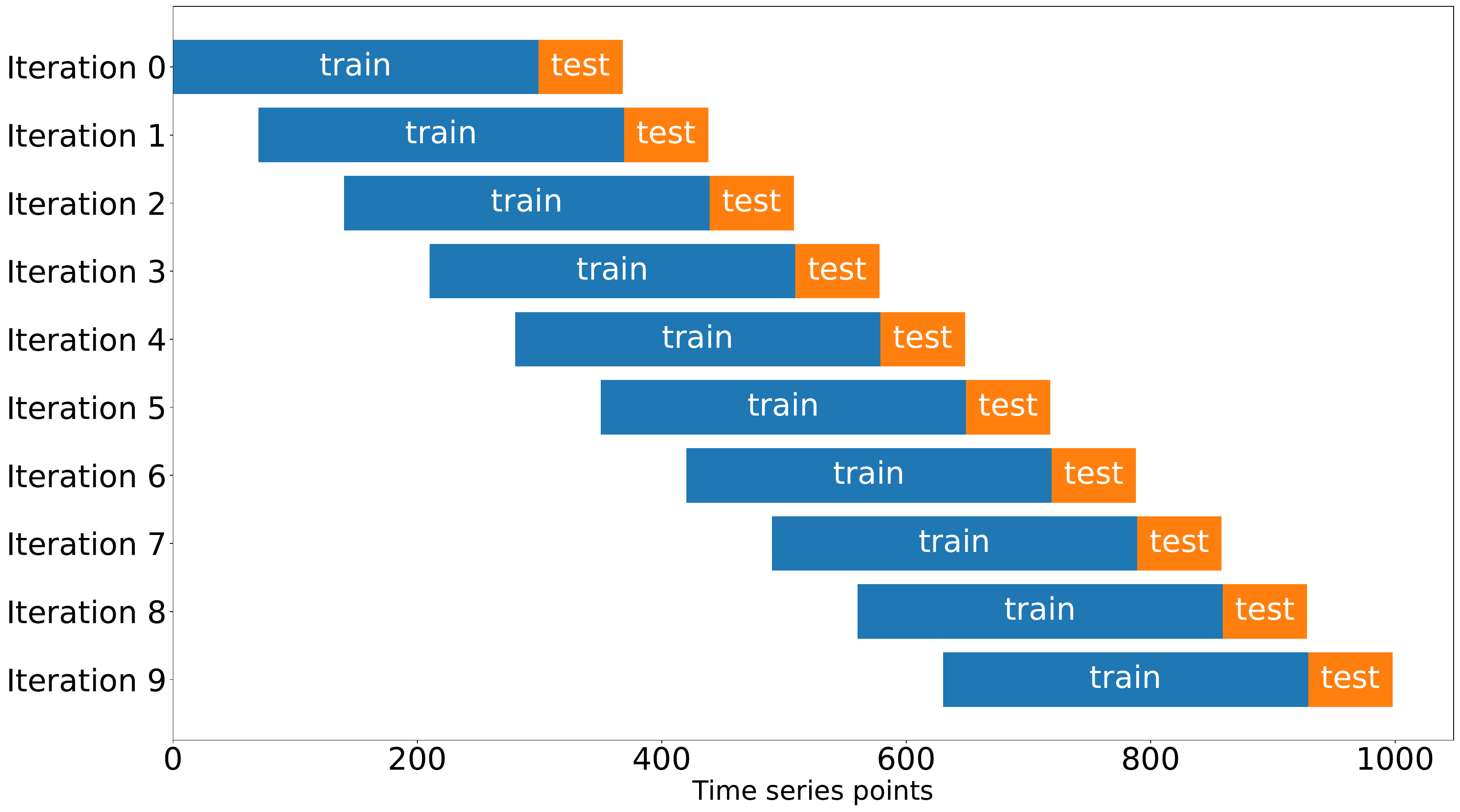}
        \caption{Data partitioning in training and test sets}
        \label{fig:timeseries_split}
    \end{figure}

    To train and test the detection model, we split $\hat{X}$ using the function TimeSeriesSplit from the scikit-learn library\footnote{\url{https://scikit-learn.org/stable/modules/generated/sklearn.model\_selection.TimeSeriesSplit.html}}. This function creates multiple training and test sets time series as illustrated in Figure \ref{fig:timeseries_split}. We split $\hat{X}$ with ten folds and fix the training/test size to $70\%-30\%$.
    
\begin{figure}[ht!]
        \centering
        
        \subfloat[Prediction and Boundaries]{
            \includegraphics[width=\linewidth]{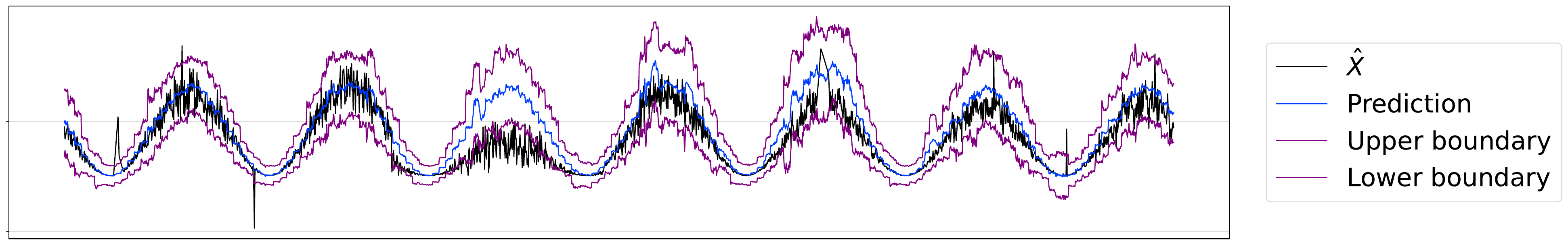}
            \label{fig:pred_boundaries}
        }
        \\
        \subfloat[Points status decision]{
            \includegraphics[width=\linewidth]{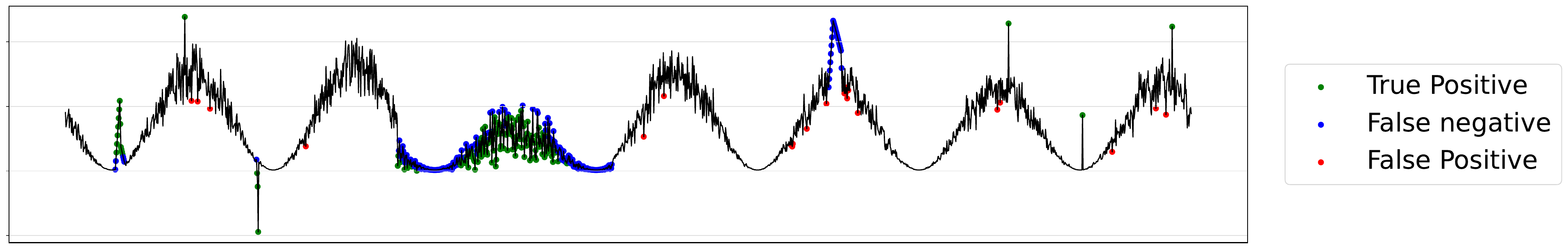}
            \label{fig:points_status}
        }
        \caption{TCN anomaly detection}
        \label{fig:tcn_detection}
    \end{figure}

    For each fold of the TimeSeriesSplit function, we apply the fitted TCN model to the test set resulting in the predicted time series $\check{X}$ and the confidence interval $\delta$. We label a point of $\hat{X}$ as anomalous if and only if:
    
    \[
        \left|\hat{X} - \check{X} \right| > \delta,
    \]
    See illustration in Figure \ref{fig:tcn_detection}.
    
    \paragraph{Analysis windows} We denote aSIM and aREAL the simulated and real analysis windows resulting from the anomaly detection model. An analysis window is a sub-sequence time series of $\hat{X}$ of fixed size $2m$, where m is the margin period. The default sampling of the time series is set to $1$ point/minute. The initial size of an analysis windows is set to $240$ points (4 hours of data centered on the anomaly) with $m$ corresponding to 2 hours of data ($120$ points). Taking a sampling period $t_s=5$ minutes (which creates a time series $X$ with a point each $5$ minutes), the margin size of the analysis windows are set to $m=24$ points. This margin size is used to center the anomaly in the analysis window. We create the analysis windows from the test set during the cross-validation. We denote $S$ the list containing the point states with the states "normal" or "anomaly". We create the analysis windows as illustrated in Algorithm \ref{alg:analysis_windows_creation}. 
    
    \RestyleAlgo{ruled}
    \begin{algorithm}
    
        \DontPrintSemicolon
        \SetKwData{Left}{left}\SetKwData{This}{this}\SetKwData{Up}{up}
        \SetKwFunction{Union}{Union}\SetKwFunction{FindCompress}{FindCompress}
        \SetKwInOut{Input}{Input}
        \SetKwInOut{Output}{Output}
        \Input{$\hat{X}$ the time series with anomalies\\$S$ the list containing the state of each point ("normal" or "anomalous")\\$m$ the analysis windows margin}
        \Output{$a$ the analysis windows list aSIM or aREAL}
        \BlankLine
        $i \gets 0$\;
        $a \gets$ empty list\;
        \While{not at the end of $S$}{
            \eIf{$S(i)$ is "anomaly"}{
                $n_c \gets$ number of consecutive points detected as "anomaly"\;
                new$_a \gets$ sub-sequence time series $\hat{X}_{[i - m, i + m]}$\;
                $a \gets a$ + new$_a$\;
                $i \gets i + n_c$\;
            }{
                $i \gets i + 1$\;
            }
        }
        \caption{Analysis windows creation}
        \label{alg:analysis_windows_creation}
    \end{algorithm}

    It should be noted that (1) a given analysis window may contain multiple anomalous points, (2) the anomalous points found in a given analysis window may be related to different anomalies, (3) analysis windows may overlap (meaning we could have the same anomaly present in different analysis windows), (4) when simulated data is used, analysis windows are defined independently from the anomaly windows produced by the simulator. 
    We evaluate our anomaly detection system by computing its F1 score from the following measures:
    \begin{itemize}
        \item True positives (TP): the number of true anomalies that are included in at least one analysis window.
        \item False positives (FP): the number of analysis windows that do not contain any true anomaly.
        \item False negatives (FN): the number of true anomalies that are not included in any analysis window.
    \end{itemize}
    This definition of the F1 score is sensitive to the size $2m$ of an analysis window. Indeed, increasing $2m$ mechanically increases the F1 score by reducing the number of false positives and false negatives, and increasing the number of true positives.

\subsection{Anomaly Classification}
\label{subsec:anomaly_classification}

    \paragraph{Time series decomposition} During the training of the detector (Section~\ref{subsec:anomaly_detection}), we also apply the additive Moving average time series decomposition\footnote{\url{https://www.statsmodels.org/dev/generated/statsmodels.tsa.seasonal.seasonal_decompose.html}} from the statsmodel library:
    \[
       \hat{X} =  T_i + U_i + R_i,
    \]
    where $\hat{X}$ is the time series containing anomalies, $T_i$ is the trend component, $U_i$ is the seasonal component, and $R_i$ is the residual. As the seasonality amplitude could vary each week in real uses, we decided to extract the seasonality from the last 2 weeks (as we need a minimum of 2 occurrences to extract the seasonality) of the train time series. We also extracted the trend from the last week of the train time series. This decomposition is done to evaluate the analysis windows labels based on the last observations of the train time series (as the detection model gives a week prediction based on the last 8 days). The seasonality and the trend extracted are subtracted to the test set. This decomposition is applied after the prediction and before the analysis windows creation. The classifiers are then trained on $R_i$. We use the Moving average decomposition because it gives a more stable seasonality compared to the STL (Seasonal and Trend decomposition using Loess~\cite{cleveland1990stl}) and more versatile than traditional decomposition methods such as X11 and SEATS (Seasonal Extraction in ARIMA Time Series) \cite{dagum2016seasonal,hyndman2018forecasting}.    
    
    \paragraph{Classification models} We tested a total of three classification models, including two classical machine-learning models, namely k-Nearest Neighbors (kNN) and Supervised Time Series Forest (STSF~\cite{random_forest}), as well as one Deep Learning model, a Temporal Convolutional Network (TCN~\cite{bai2018empirical}).
    
    We used the kNN implementation of the sktime library\footnote{\url{https://www.sktime.org/en/stable/api\_reference/auto\_generated/sktime.classification.distance\_based.KNeighborsTimeSeriesClassifier.html}}, varying the parameters according to Table~\ref{table:knn}. We use the k-Nearest neighbor as it is one of the standard classification method used for time series. One of the combination used in Table~\ref{table:knn}, 1-NN with the Dynamic Time Warping distance (DTW) is often used in time series analysis \cite{susto2018time}.
    STSF is an ensemble of decision trees that samples the training set with class-balanced bagging and creates intervals for 3 representations (the time series, the frequency domain and the derivative representation of the time series) and 7 features (mean, median, standard deviation, slope aggregation functions, inter-quartile range, minimum and maximum). We used the STSF implementation of the sktime library\footnote{\url{https://www.sktime.org/en/stable/api\_reference/auto\_generated/sktime.classification.interval\_based.SupervisedTimeSeriesForest.html}} and we varied the number of estimators between 5 and 200. We use this interval-based classifier for its efficiency and also because it is an enhanced method of Random Forest traditionally used in time series classification.
    STSF is an ensemble of decision trees that samples the training set with class-balanced bagging and creates intervals for 3 representations (the time series, the frequency domain and the derivative representation of the time series) and 7 features (mean, median, standard deviation, slope aggregation functions, inter-quartile range, minimum and maximum). We used the STSF implementation of the sktime library\footnote{\url{https://www.sktime.org/en/stable/api\_reference/auto\_generated/sktime.classification.interval\_based.SupervisedTimeSeriesForest.html}} and we varied the number of estimators between 5 and 200. We use this interval-based classifier for its efficiency and also because it is an enhanced method of Random Forest traditionally used in time series classification.
    For the Temporal Convolutional network, we used the Keras TCN implementation\footnote{\url{https://github.com/philipperemy/keras-tcn}} described in~\cite{bai2018empirical}, varying the parameters according to Table~\ref{table:tcn}. The TCN model used a softmax activation function for its final layer and the sparse categorical cross-entropy loss function. We use the TCN method as it achieves good performance compared to Recurrent Neural Network (RNN) and allows parallel computation for the outputs. We wanted to evaluate the TCN performance for classification tasks in time series as it is mainly used in anomaly detection problems. 

    \begin{table}[ht!]
        \centering
        \begin{tabular}{|l|l|}
            \hline
            Parameter & Values tested\\
            \hline
            Number of neighbours & 1, 3, 5, 10, 20, 50\\\hline
            Distance & \makecell[l]{dtw, ddtw, wdtw, lcss, erp, msm,\\ twe}\\\hline
            Weight & uniform, distance\\
            \hline
        \end{tabular}
        \caption{Parameter combinations tested for the kNN time series classifier. Dtw: dynamic time warping, ddtw: derivative dtw, wdtm: weighted dtw, lcss: longest common sub-sequence, erp: edit distance with real penalty, msm: move split merge, twe: time warping edit}
        \label{table:knn}
    \end{table}

    \begin{table}[ht!]
        \centering
        \begin{tabular}{|l|l|}
            \hline
            Parameter & Values tested\\\hline
            Kernel size & 4\\\hline
            Activation function & softsign, relu, tanh\\\hline
            Number of filters & 64\\\hline
            Optimizer & \makecell[l]{RMSprop, Adam, Adamax, Nadam}\\\hline
        \end{tabular}
        \caption{Parameter combinations tested for the TCN time series parameter. RMSprop: root mean square propagation}
        \label{table:tcn}
    \end{table}
    
    To evaluate the classifiers, we split the analysis windows into a training and test sets with a proportion of $70\%-30\%$. We applied the Grid search cross validation\footnote{\url{https://scikit-learn.org/stable/modules/generated/sklearn.model\_selection.GridSearchCV.html}} and the Stratified k-Fold\footnote{\url{https://scikit-learn.org/stable/modules/generated/sklearn.model\_selection.StratifiedKFold.html}} iterator from the scikit-learn library to search for the optimal parameters on the training set. The stratify strategy keeps the same class proportion between the folds and the data set. We evaluated the models using the canonical micro F1 score definition and confusion matrix for each anomaly class.

    \section{Experiments and Results}
    \label{sec:experiments}
	We conducted experiments to evaluate our anomaly classification method using simulated and real data sets. We also evaluated the performance of the detection method as a sanity check since the performance of the classification model clearly depends on the type of detected anomalies. 

\subsection{REAL and aREAL data sets preparation}
\label{subsec:real_preparation}

    \paragraph{REAL time series} Before using the REAL time series, we cleaned the missing parts of the time series (where $\hat{X}(t)=0$) that usually result from interruption of the monitoring system and are therefore excluded from the anomaly detection model. To clean these time series, we simply deleted the missing parts from the time series. As the TCN needs a continuous time series to fit the model, we rebuilt the missing parts by extracting the weekly seasonality component using STL from the rest of the time series. The missing parts were filled with the mean of $o$ points occurring at the same weekly periodicity time (for example each Monday), where $o$ is an integer between $1$ and $3$ depending on the size of the sane parts (the maximum of sane occurrence up to $3$). We used the STL decomposition for the seasonality extraction as the seasonal component evolve over time compared to the moving average method. We discarded the time series with more than $10\%$ missing parts. We also applied the same sampling period $t_s$ to the REAL time series as well as the filtering and scaling processing in subsection~\ref{subsubsec:data_preparation_noise_injection}.  
    
    We also measured the noise level from each REAL set using a high-pass Butterworth filter with the following parameters:
    \begin{alignat*}{3}
        N & = 5  &\qquad  f_s & = \dfrac{1}{t_s \times 60} &\qquad  W_c & = \dfrac{f_s}{8},
    \end{alignat*}
    where $N$ is the filter order, $f_s$ is the sampling frequency, and $W_c$ the critical frequency and $t_s$ the sampling period defined in Section~\ref{subsec:simulation}. After extracting the high frequencies, we calculated the noise level as the standard deviation of the time series with a degree of freedom equal to $1$. The noise level is needed during the experiments to compare results obtained from the same noise level.
    
\begin{figure}[ht!]
        \centering
        \includegraphics[width=\linewidth]{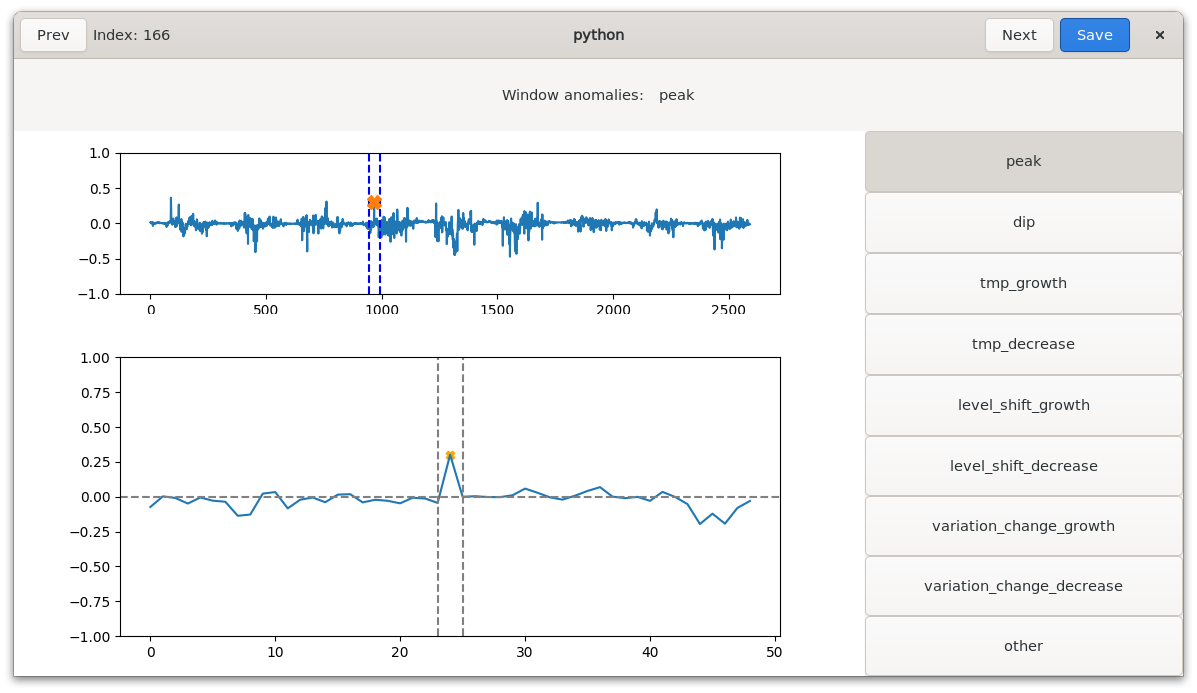}
        \caption{Manual classification program}
        \label{fig:manual_classification}
    \end{figure}

    \paragraph{aREAL labeling} The analysis windows extracted from the real dataset were manually labeled using the classification program shown in Figure~\ref{fig:manual_classification}, where the user can assign one or multiple labels for each anomaly detected. The last label ``other" is used in case the anomaly cannot be clearly classified into existing classes. The proportion of anomalies classified as ``other" represented $0.8\%$ of the total number of analysis windows. 
    
\subsection{Datasets}
\label{subsec:datasets}

    \paragraph{REAL} The first data set is composed of the REAL time series with $65$ different 1-month-long time series. The $65$ time series correspond to the average delay collected from  $65$ different networks labeled REAL1 to REAL65. Each of these sources has a length of 1 month and may have a different amplitude and seasonality.
    
    \paragraph{aREAL} The second data set consists of the aREAL analysis windows created by the Anomaly detection model on the REAL time series. We named the analysis windows from aREAL1 to aREAL3 depending on the range of their noise level (see Table~\ref{table:analysis_windows_parameter}) as explained previously. We assigned class labels to the analysis windows manually, using the program described previously in Section~\ref{subsec:real_preparation}, resulting in the class proportions reported in Table~\ref{table:anomalies_weights}. Most of the detected anomalies were Single point or Temporary change anomalies with the peak or growth direction.
    
    \begin{table}[ht!]
        \centering
    
        \begin{tabular}{|l|c|}
            \hline
            Dataset & Range of noise level $\sigma$ \\\hline
            aSIM1 & $[0.0, 0.01[$\\\hline
            aSIM2 and aREAL1 & $[0.01, 0.03[$\\\hline
            aSIM3 and aREAL2 & $[0.03, 0.05[$\\\hline
            aSIM4 and aREAL3 & $[0.05, 0.07[$\\\hline
            aSIM5 & $[0.07, 0.09[$\\\hline
        \end{tabular}
    
        \caption{Parameter combinations used for the aSIM or aREAL separation}
        \label{table:analysis_windows_parameter}
        
    \end{table}

    \begin{table}[ht!]
        \centering
    
        \begin{tabular}{|l|c|c|c|}
            \hline
            \multirow{3}{*}{Anomaly class} & \multicolumn{3}{c|}{Subclasses and class proportion}\\\cline{2-4}
            & Peak or & Dip or & \multirow{2}{*}{Total}\\
            & Growth & Decrease &\\\hline
            Single point & 0.43 & 0.02 & 0.45\\\hline
            Temporary Change & 0.38 & 0.02 & 0.4\\\hline
            Level Shift & 0.005 & 0.005 & 0.01\\\hline
            Variation Change & 0.1 & 0.04 & 0.14 \\\hline
        \end{tabular}
        \caption{Proportion of assigned anomaly types in the aREAL dataset}
        \label{table:anomalies_weights}
    \end{table}
    
    \paragraph{SIM} The third data set is the SIM time series generated as detailed in Section~\ref{subsec:simulation}. We extracted the noise level from the REAL time series and the proportion of each class of anomalies from the aREAL analysis windows, and injected them in the simulator. We generated a total of 54 data sets, varying the parameters according to Table~\ref{table:simulated_parameters}.
    
    \paragraph{aSIM} The fourth data set is composed of the aSIM analysis windows produced by the anomaly detection model on the SIM time series, named aSIM1 to aSIM5 depending on their noise level (Table~\ref{table:analysis_windows_parameter}).
    
    \begin{table}[ht!]
        \centering
        \begin{tabular}{|l|l|}
            \hline
            Parameter & Values\\
            \hline
            Number of anomalies $n$ & 250, 500, 1000\\\hline
            Noise level $\sigma$ & 0.0, 0.2, 0.4, 0.6, 0.8\\\hline
            Strength $\alpha$ & [0.5, 0.7]\\\hline
            Anomaly proportion  & \makecell[l]{Imbalanced: (0.43, 0.02,\\ 0.38, 0.02,\\ 0.005, 0.005,\\ 0.1, 0.04),\\ Balanced: (0.125, 0.125,\\ 0.125, 0.125,\\ 0.125, 0.125,\\ 0.125, 0.125)}\\
            \hline
        \end{tabular}
    
        \caption{Parameter combinations used for the SIM dataset creation}
        \label{table:simulated_parameters}
        
    \end{table}

\subsection{Training and evaluation}
\label{subsec:training_evaluation}

    As explained previously, an anomaly can appear in multiple analysis windows and therefore the anomaly proportion may not be the same between the SIM and aSIM sets. For our experiments, we wanted to keep the same proportion between the detection and classification as the imbalanced proportion case is extracted from the aREAL analysis windows. We used two methods to guarantee the same proportion between the anomalies in the SIM and aSIM data sets. In case of a balanced proportion, we downsampled the majority class using the RandomUnderSampler\footnote{\url{https://imbalanced-learn.org/stable/references/generated/imblearn.under\_sampling.RandomUnderSampler.html}} function from the Imbalanced-learn library. In case of an imbalanced proportion, we randomly selected samples from each class according to the anomaly proportion parameter used in the Anomaly Simulation.

    \paragraph{SIM-SIM} We named the first experiment SIM-SIM as it trained and evaluated the detection and classifiers on the SIM and aSIM data sets, having prepared the aSIM analysis windows to respect the class proportions as detailed before. The goal of this experiment was to evaluate the detection and classification parts on the simulated data sets. In the second experiment, we compared these results to the ones obtained with the REAL data set.
    
    \paragraph{SIM-REAL} We named the second experiment SIM-REAL as it trained the classifier on the aSIM dataset and evaluated it on the aREAL dataset. We trained different classifiers for each range of noise level defined in Table~\ref{table:analysis_windows_parameter}. 
    and we evaluated them independently. The data sets used are aSIM2-4 and aREAL1-3.

\subsection{SIM-SIM results}
\label{subsec:sim_sim}

    \subsubsection{Detection results}
    \label{subsubsec:detection_results}
    
        \begin{figure}[H]
            \centering
            \subfloat[Imbalanced data sets\label{fig:sim_imbalanced}]{
                \includegraphics[width=\linewidth]{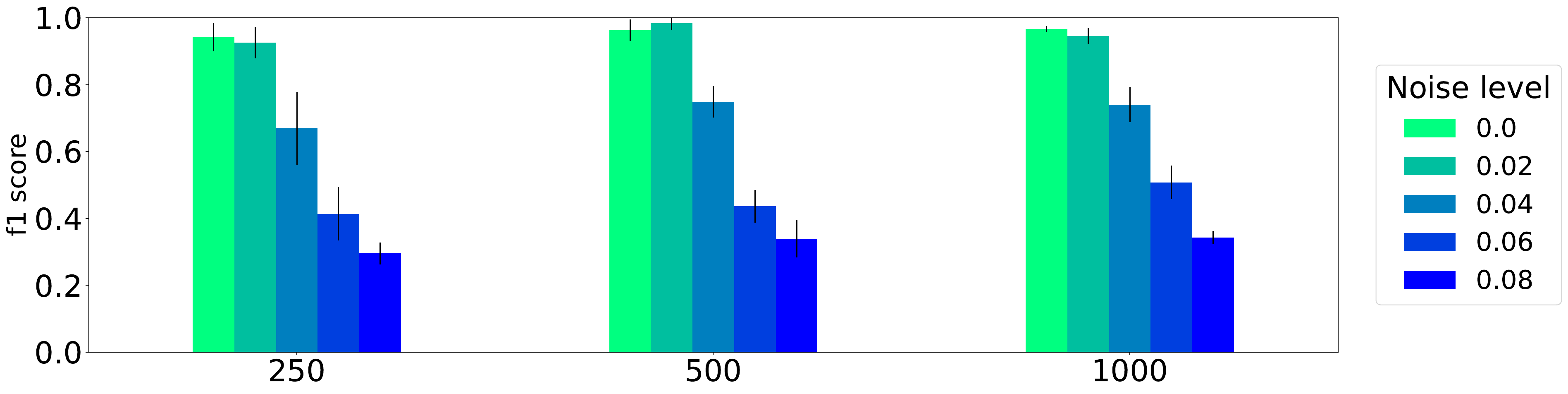}
            }
            \\
            \subfloat[Balanced data sets\label{fig:sim_balanced}]{
                \includegraphics[width=\linewidth]{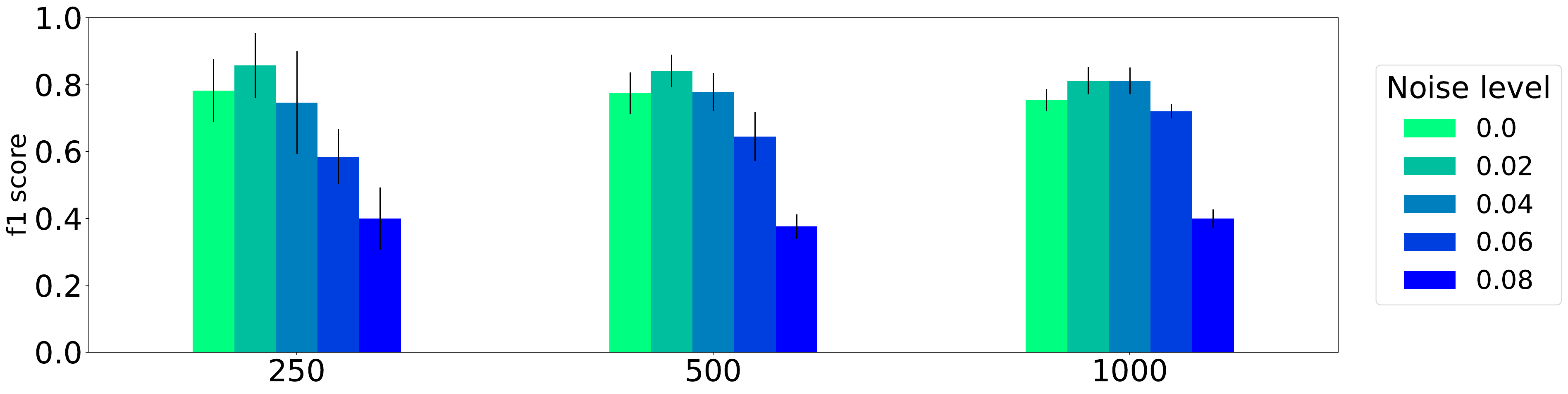}
            }
            \caption{Detection results, SIM-SIM datasets ($t_s=5$ minutes)}
            \label{fig:sim}
        \end{figure}      
    
        Figure~\ref{fig:sim} shows the performance of the detection model for different noise levels. For the imbalanced case, the F1 score decreased as the noise level increased in the SIM dataset $\tilde{ts}$. The balanced case showed an increase until $\sigma=0.1\mu s$ and then a rapid decrease. The difference in behavior between the imbalanced and balanced dataset reflects the detection of the level shift and variation change classes. These classes are more difficult to detect than the other ones, which explains the reduced performance in the balanced dataset. In the imbalanced dataset, these classes are too infrequent to have a measurable impact on the performance as we used the micro F1 score.
    
    \subsubsection{Classification results}
    \label{subsubsec:classification_results}
        
\begin{figure*}
            \includegraphics[width=\linewidth]{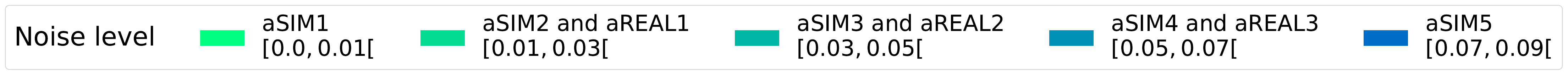}
            \subfloat[All anomaly types]{
                 \includegraphics[width=0.40\linewidth]{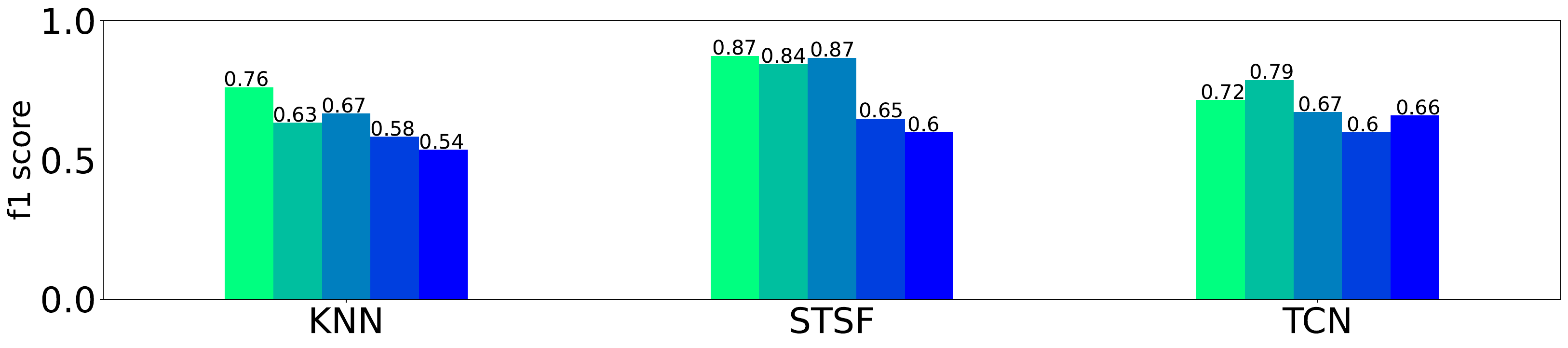}
                }
            \subfloat[STSF classifier]{
                \includegraphics[width=0.40\linewidth]{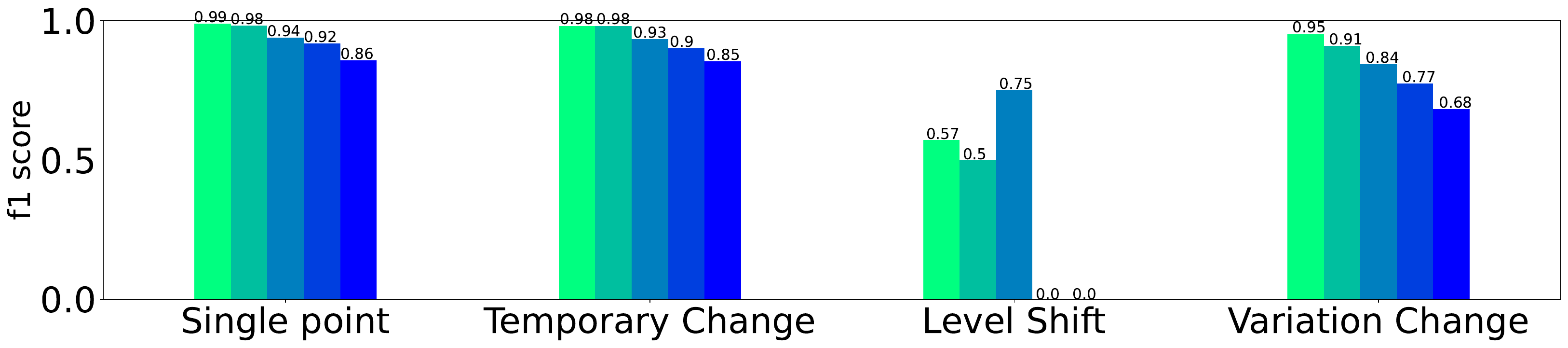}
            }\\
            \subfloat[TCN classifier]{
                \includegraphics[width=0.40\linewidth]{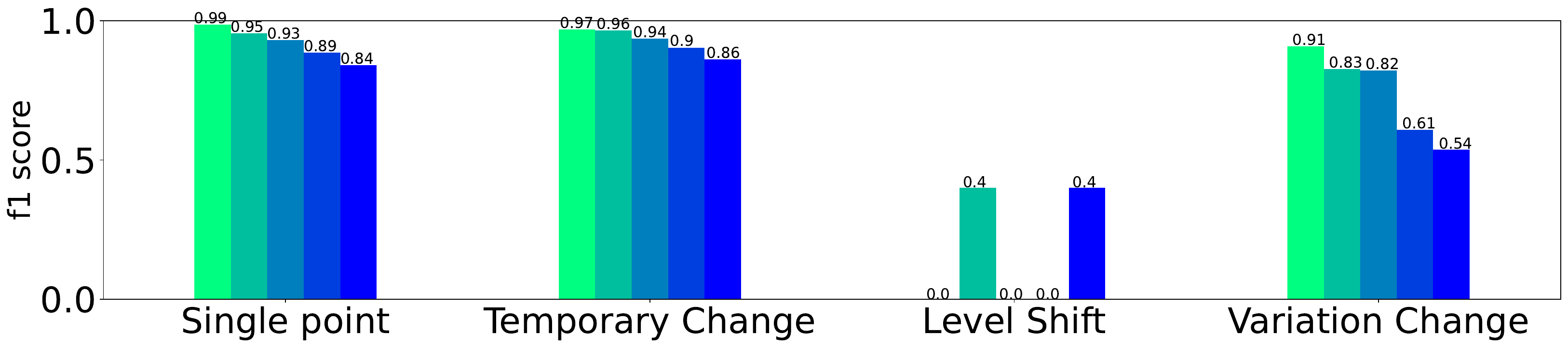}
            }
            \subfloat[kNN classifier]{\includegraphics[width=0.40\linewidth]{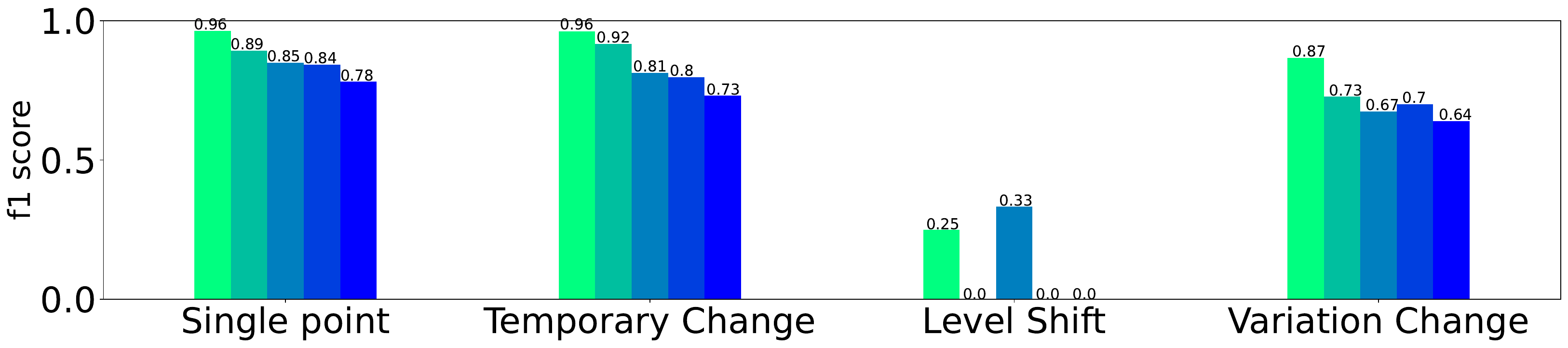}}
            \caption{SIM-SIM classification results (imbalanced data)}
            \label{fig:sim_sim_classification_imbalanced}
        \end{figure*}

\begin{figure*}
            \subfloat[All anomaly types]{
                 \includegraphics[width=0.40\linewidth]{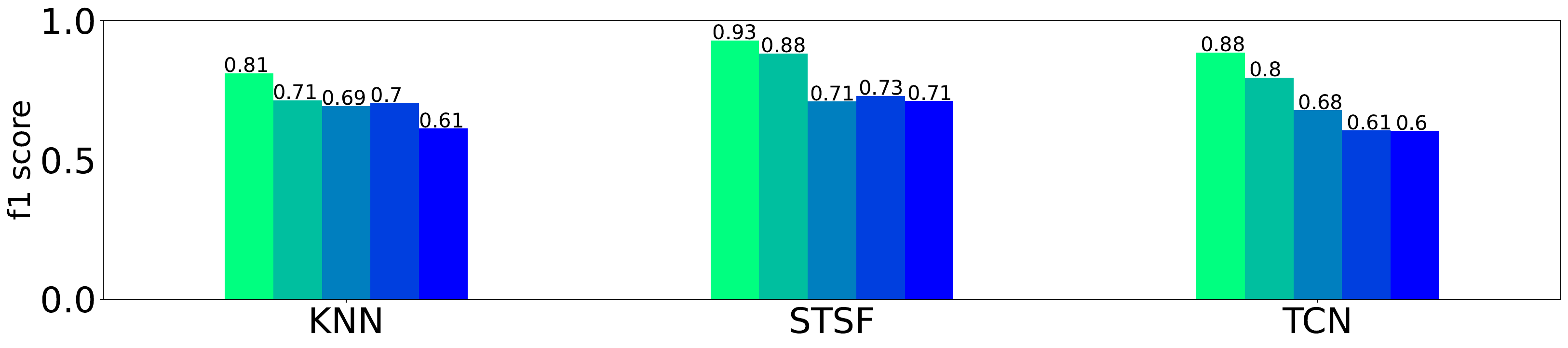}
                }
            \subfloat[STSF classifier]{
                \includegraphics[width=0.40\linewidth]{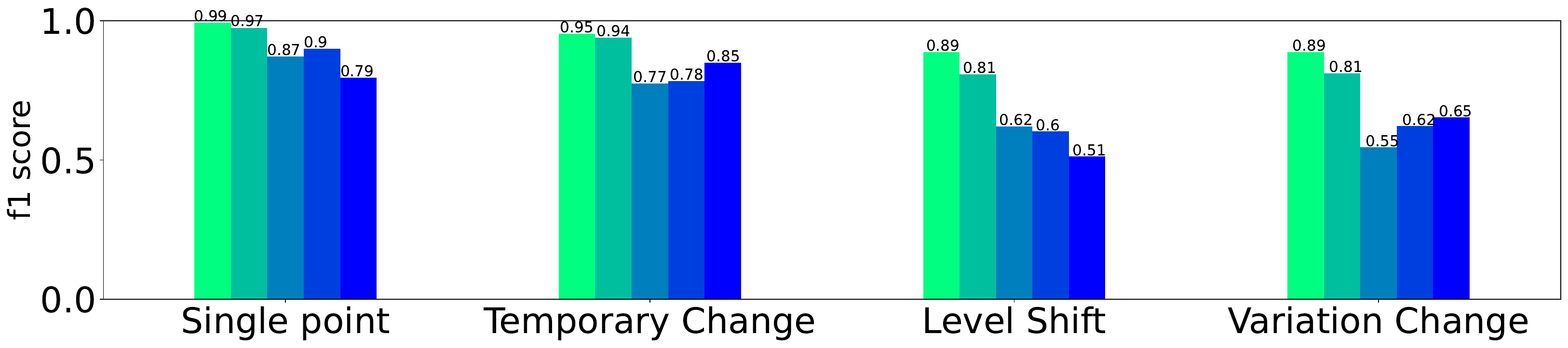}
            }\\
            \subfloat[TCN classifier]{
                \includegraphics[width=0.40\linewidth]{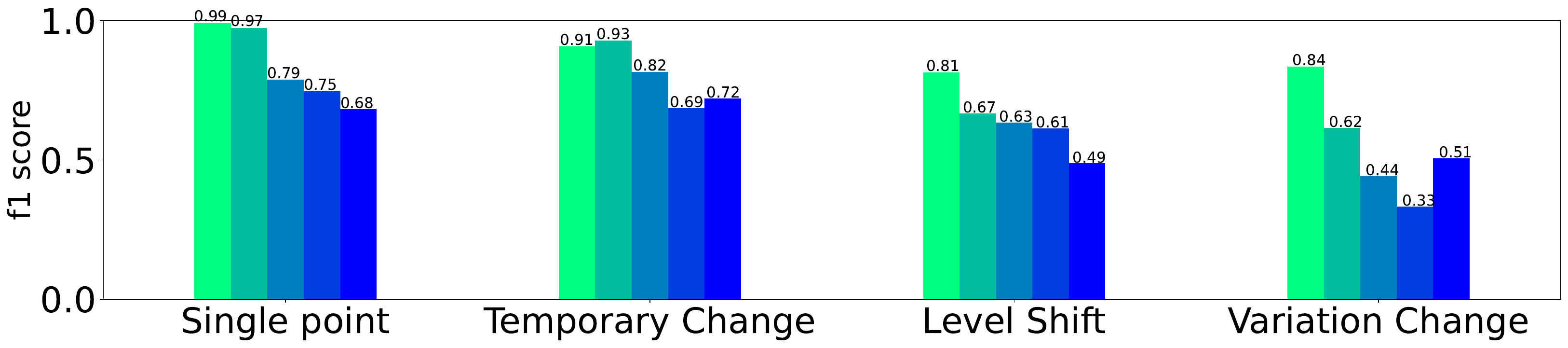}
            }
            \subfloat[kNN classifier]{\includegraphics[width=0.40\linewidth]{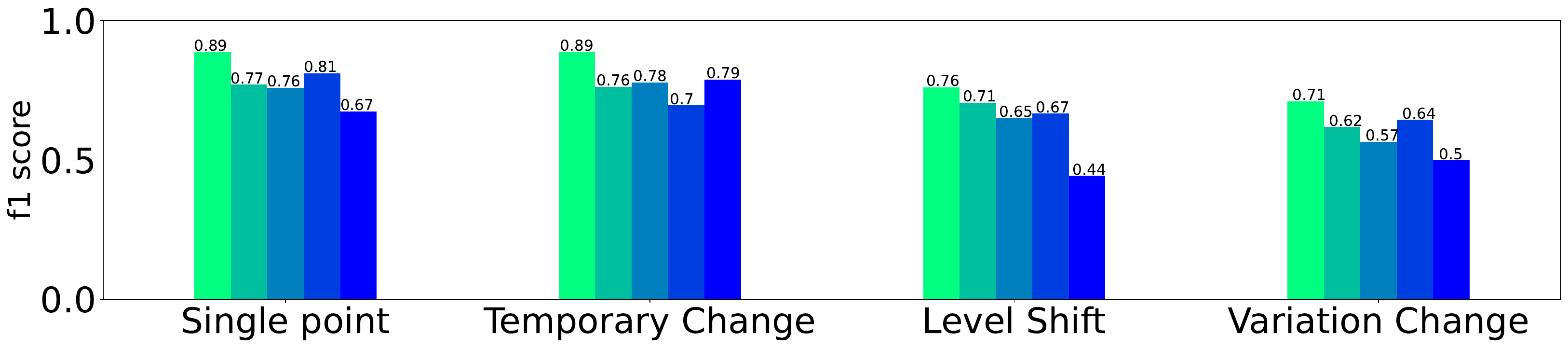}}
            \caption{SIM-SIM classification results (balanced data)}
            \label{fig:sim_sim_classification_balanced}
        \end{figure*}

        Figures~\ref{fig:sim_sim_classification_imbalanced}, \ref{fig:sim_sim_classification_balanced} show an overview of the classifiers results for different noise levels and weights. The STSF and TCN models perform the best in both the imbalanced and balanced case. The overall F1 score of the classifiers across all anomaly classes is equal to $0.7$ for the imbalanced case and $0.73$ for the balanced case. As expected, the F1 score decreases as the noise level increases. We evaluated a decrease of the F1 score around $29\%$ for the TCN and around $23\%$ for the STSF between low (aSIM1) and high noise level (aSIM5). The Single point and Temporary change anomalies have the best overall F1 score.

        The detection and classification performs well for the simulated data sets. The detection shows good results for both the imbalanced and balanced case with a F1 score higher than chance even for high level of noise.

\subsection{SIM-REAL results}
\label{subsec:sim_real}
    
\begin{figure*}
        \subfloat[All anomaly types]{
             \includegraphics[width=0.40\linewidth]{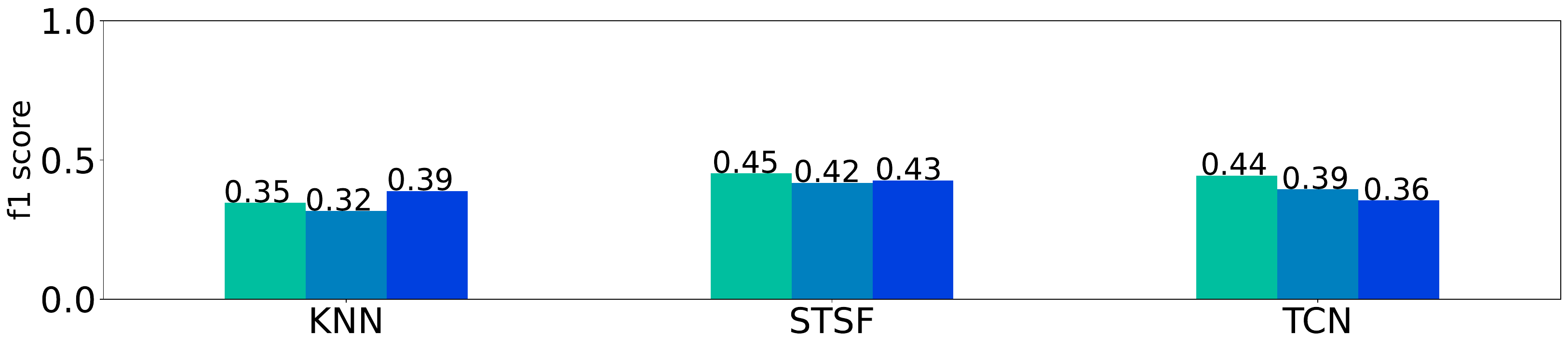}
            }
        \subfloat[STSF classifier]{
            \includegraphics[width=0.40\linewidth]{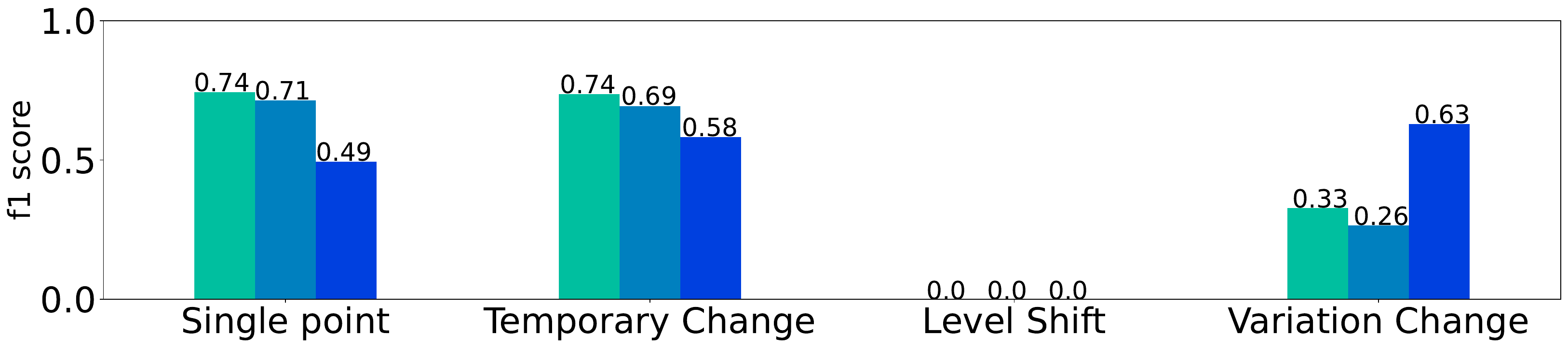}
        }\\
        \subfloat[TCN classifier]{
            \includegraphics[width=0.40\linewidth]{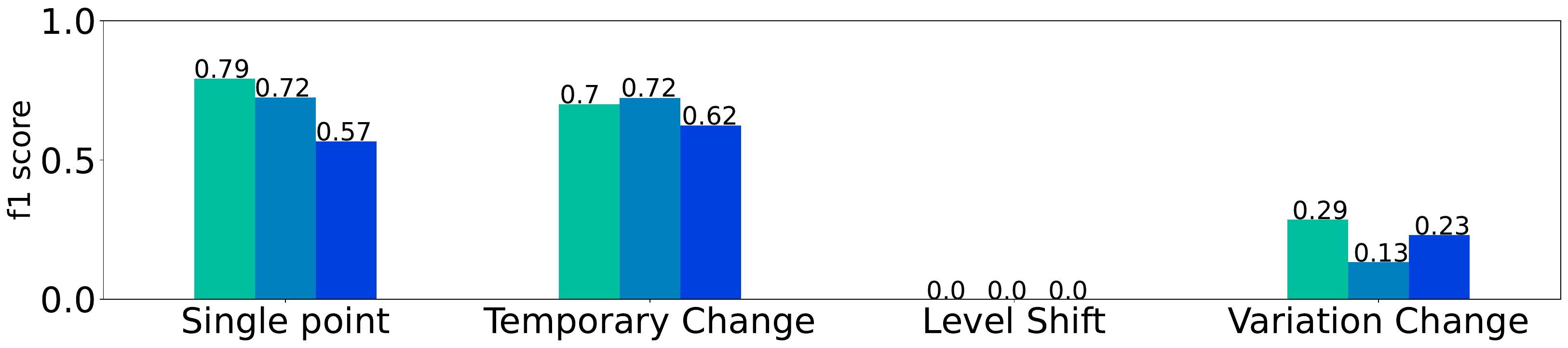}
        }
        \subfloat[kNN classifier]{\includegraphics[width=0.40\linewidth]{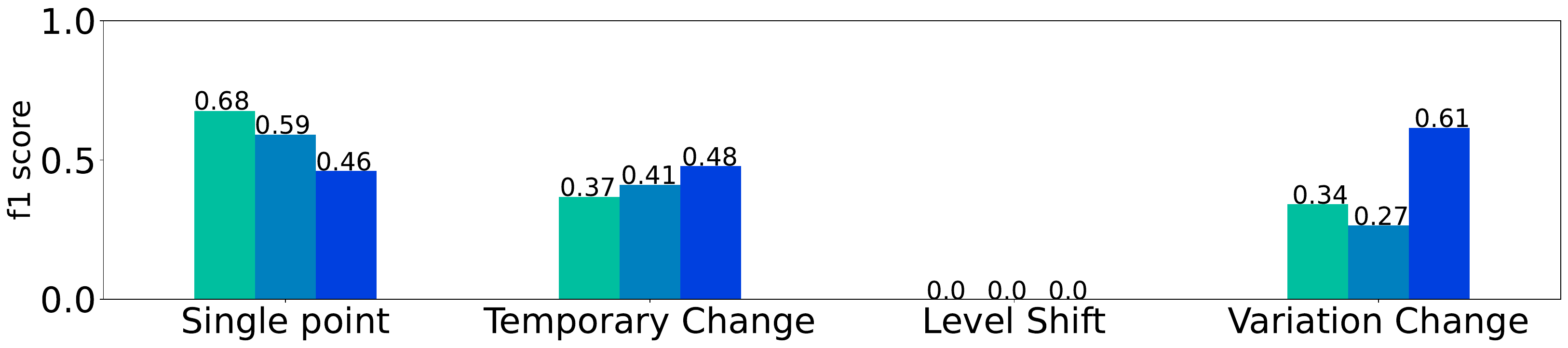}}
        \caption{SIM-REAL classification results (imbalanced data)}
        \label{fig:sim_real_classification_imbalanced}
    \end{figure*}

\begin{figure*}
        \subfloat[All anomaly types]{
             \includegraphics[width=0.40\linewidth]{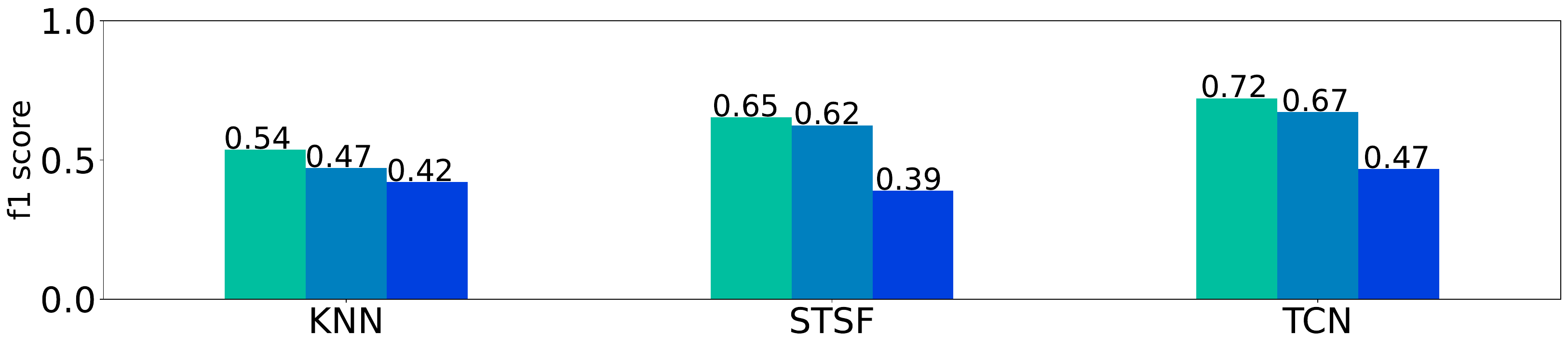}
            }
        \subfloat[STSF classifier]{
            \includegraphics[width=0.40\linewidth]{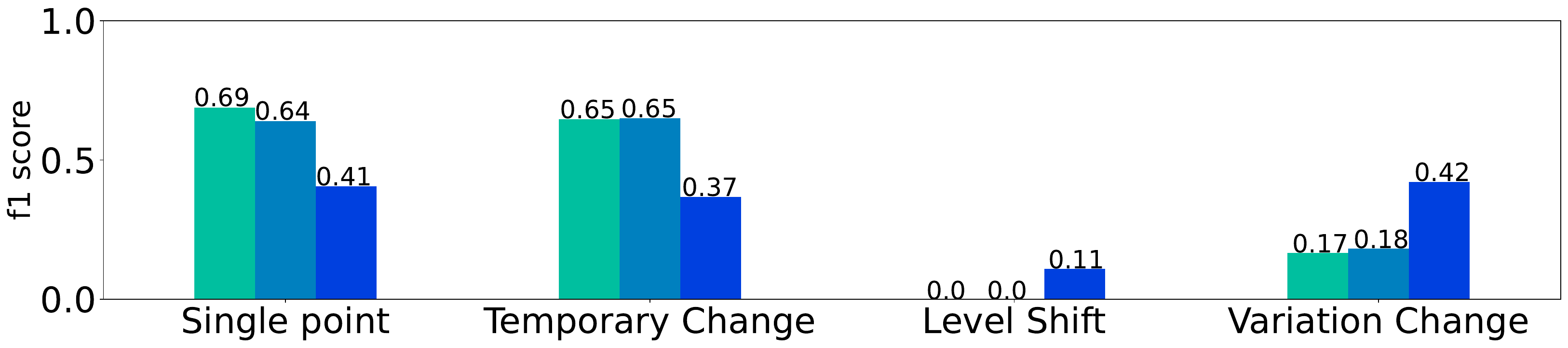}
        }\\
        \subfloat[TCN classifier]{
            \includegraphics[width=0.40\linewidth]{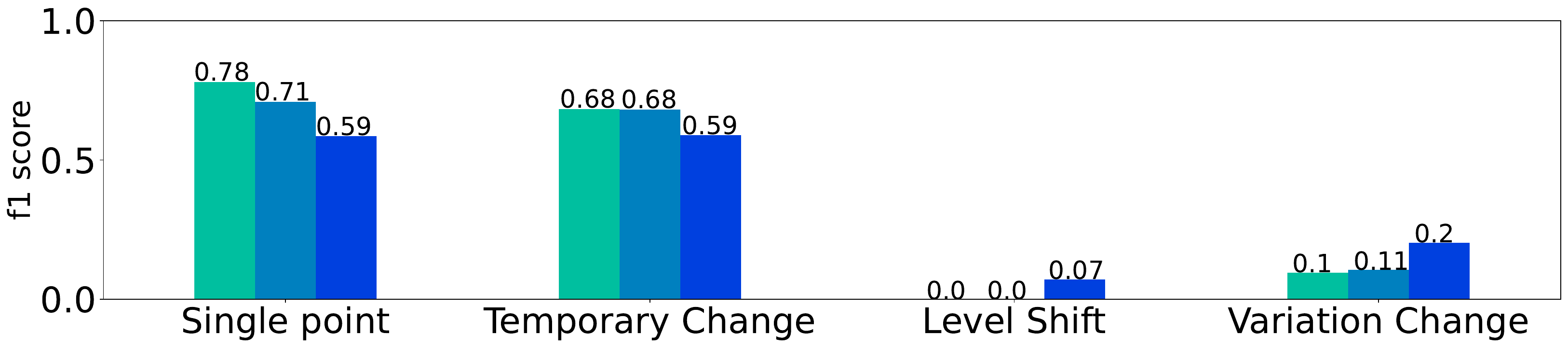}
        }
        \subfloat[kNN classifier]{\includegraphics[width=0.40\linewidth]{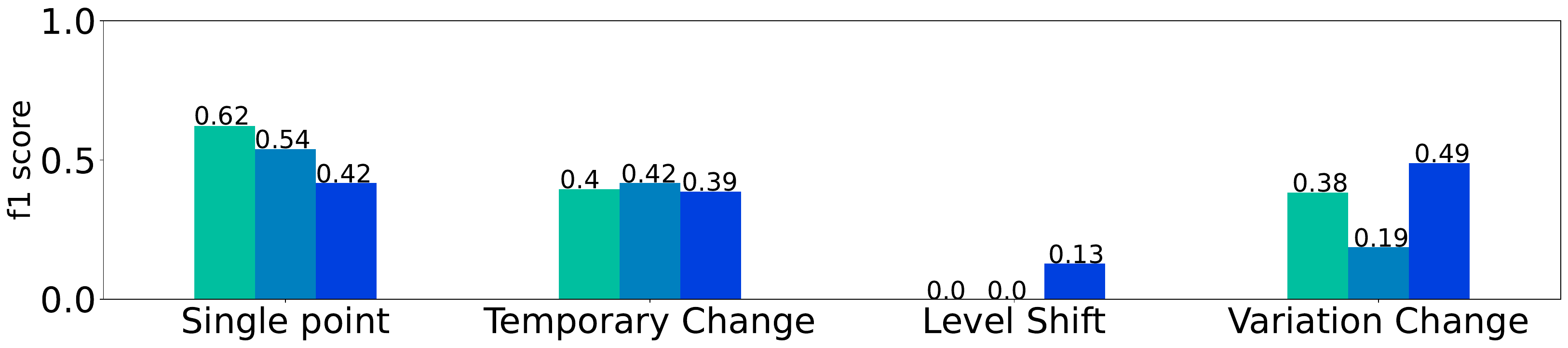}}
        \caption{SIM-REAL classification results (balanced data)}
        \label{fig:sim_real_classification_balanced}
    \end{figure*}

    As shown in Figure~\ref{fig:sim_real_classification_imbalanced}, \ref{fig:sim_real_classification_balanced}, the STSF and TCN  classifiers show good performances for Single point and Temporary change anomalies, even though they were only trained on the simulated dataset. The kNN classifier did not perform as well as the STSF and TCN classifiers. The overall F1 score of the classifiers decreases to 0.39 for the imbalanced case and 0.55 for the balanced case considering all the anomaly types. This relatively poor performance is explained by the difficulty to classify level shifts and variation changes. The overall F1 decrease is due to the insufficient number of anomalies for the Level shift and Variation change types. This type of anomalies are not frequent in the REAL data sets observed. We evaluated a decrease of the F1 score around $24\%$ for the TCN and around $35\%$ for the STSF between the simulated and real-world datasets.

	\section{Conclusion and Future work}
    \label{sec:conclusion}
    The classifiers trained on simulated data (aSIM dataset) achieve excellent performance (F1 in [0.6-0.93] depending on noise level) when tested on simulated data, which demonstrates the relevance of the approach. When tested on the aREAL datasets, the TCN and STSF classifiers achieve a relatively good F1 score (F1 $>$ 0.6) for the single point and temporary change anomalies and for low levels of noise. Under these assumptions, the classifiers appear to be usable in real conditions, which is remarkable given that the classifiers were only trained on simulated data. 

The poor classification of level shift and variation change anomalies results from the difficulty to separate these classes on 2-hour anomaly windows, which was observed in the simulated dataset and confirmed in the real dataset. To address this issue, one could adopt larger or adaptive anomaly window sizes.  As expected, the noise level also directly impacts the classifiers performance. Pre-filtering the signal might help reducing sensitivity to noise.

Multiple improvements could be envisaged in the future. First, the anomaly simulator could support superimposed anomalies that occur frequently in real time series. The simulator could also modulate the signal amplitude according to variation patterns observed in the real data, such as variations between week days and weekends. The detection and data preparation steps also have a large impact on the classification. 
A better integration between data preparation, anomaly detection, and anomaly classification could help improve the performance of the classifier. For instance, we could create a detector for each class of anomalies and optimize their parameters according to it (for example an hour prediction for the Single point anomalies). Another integration could be to give more information about the anomalies such as the approximate length. The length could be calculated depending the length of the time series which contains around 80-90\% or anomalous points.

	\section*{Acknowledgment}
	This work was funded by Mitacs (\url{https://www.mitacs.ca}) under award number IT16266.
	
	\ifCLASSOPTIONcaptionsoff
	\newpage
	\fi

	
	
	%
	
	\bibliographystyle{IEEEtran}
	\bibliography{IEEEabrv,Biblio/Anomaly}
	
	%
	
	
	
	
	
	
	
	
	
\end{document}